\documentclass{article}

\usepackage{microtype}
\usepackage{graphicx}
\usepackage{booktabs}
\usepackage[title]{appendix}

\usepackage{hyperref}

\usepackage{algorithm}
\usepackage{amsmath}
\usepackage{enumitem}
\usepackage{pgfplots}
\usepackage{tikz}
\usepackage{xspace}
\usepackage{subcaption}

\newcommand{\ourtitle}{A Study on the Intersection of GPU Utilization\\and CNN Inference}
\newcommand{\fulltitle}{A Study on the Intersection of GPU Utilization and CNN Inference}

\newenvironment{denseenum}{
\begin{enumerate}[topsep=2pt, partopsep=0pt, leftmargin=1.5em]
  \setlength{\itemsep}{2pt}
  \setlength{\parsep}{0pt}
}{\end{enumerate}}

\newcommand{\eg}{e.g.,\xspace}
\newcommand{\ie}{i.e.,\xspace}
\newcommand{\Section}{\S}
\newcommand{\Figure}{Figure~}
\newcommand{\Equation}{Equation~}
\newcommand{\Algorithm}{Algorithm~}

\newcommand{\NATSbench}{NATS-Bench\xspace}

\newcommand{\pf}{Pareto frontier\xspace}

\newcommand{\pfs}{Pareto frontiers\xspace}

\newcommand{\nn}{NN\xspace}

\newcommand{\nnFull}{neural network\xspace}
\newcommand{\nnsFull}{neural networks\xspace}

\newcommand{\cnn}{CNN\xspace}
\newcommand{\cnns}{CNNs\xspace}

\newcommand{\ai}{arithmetic intensity\xspace}
\newcommand{\ais}{arithmetic intensities\xspace}

\newcommand{\cmr}{CMR\xspace}

\newcommand{\cmrFull}{compute-to-memory-bandwidth ratio\xspace}

\newcommand{\FLOPS}{FLOPs/sec.\xspace}
\newcommand{\op}{FLOP\xspace}
\newcommand{\ops}{FLOPs\xspace}
\newcommand{\FLOP}{FLOP\xspace}
\newcommand{\FLOPs}{FLOPs\xspace}

\newcommand{\symOn}{X\xspace}
\newcommand{\symOff}{-\xspace}

\newcommand{\searchspace}{search space\xspace}
\newcommand{\searchspaces}{search spaces\xspace}

\newcommand{\throughput}{application-level throughput\xspace}

\newcommand{\nas}{NAS\xspace}

\newcommand{\nasFull}{neural architecture search\xspace}

\newcommand{\TFLOPS}{TFLOPs/sec.\xspace}

\newcommand{\hardwareutilization}{GPU utilization\xspace}
\newcommand{\Hardwareutilization}{GPU utilization\xspace}

\newcommand{\hardwareutilizations}{GPU utilizations\xspace}

\newcommand{\timm}{\texttt{timm}\xspace}

\newcommand{\symBatchSize}{N\xspace}
\newcommand{\symHeight}{H\xspace}
\newcommand{\symWidth}{W\xspace}
\newcommand{\symChannelsIn}{C\xspace}

\newcommand{\symSearchSpace}{\mathcal{S}\xspace}
\newcommand{\symSample}{s\xspace}
\newcommand{\symEvalSet}{\mathcal{F}\xspace}
\newcommand{\symEval}{f\xspace}
\newcommand{\symEvalI}{\symEval_i\xspace}

\newcommand{\symTime}{T\xspace}
\newcommand{\symTimeBudget}{\symTime_B\xspace}
\newcommand{\symEvalTime}{t\xspace}
\newcommand{\symEvalTimeI}{\symEvalTime_i\xspace}
\newcommand{\symAcc}{A\xspace}
\newcommand{\symTput}{T\xspace}
\newcommand{\symProxy}{P\xspace}
\newcommand{\symEvalFnAcc}{\symEval_{\symAcc}\xspace}
\newcommand{\symEvalFnTput}{\symEval_{\symTput}\xspace}
\newcommand{\symEvalFnProxy}{\symEval_{\symProxy}\xspace}
\newcommand{\symEvalTimeAcc}{\symEvalTime_{\symAcc}\xspace}
\newcommand{\symEvalTimeTput}{\symEvalTime_{\symTput}\xspace}
\newcommand{\symEvalTimeProxy}{\symEvalTime_{\symProxy}\xspace}
\newcommand{\symRank}{r\xspace}
\newcommand{\symGoal}{G\xspace}
\newcommand{\symTputGoal}{\symGoal_{\symTput}\xspace}
\newcommand{\symExpTput}{w\xspace}
\newcommand{\symEvalResult}{e\xspace}
\newcommand{\symEvalResultAcc}{\symEvalResult_{\symAcc}\xspace}
\newcommand{\symEvalResultTput}{\symEvalResult_{\symTput}\xspace}
\newcommand{\symFrontier}{\mathcal{P}\xspace}
\newcommand{\symFrontierProxy}{\symFrontier_{\symProxy}\xspace}
\newcommand{\symFrontierAcc}{\symFrontier_{\symAcc}\xspace}
\newcommand{\symFrontierAccApprox}{\widehat{\symFrontierAcc}\xspace}

\usepackage[accepted]{mlsys2022}

\usepackage{fancyhdr}
\mlsystitlerunning{\fulltitle}

\begin{document}

\twocolumn[
\mlsystitle{\ourtitle}

\mlsyssetsymbol{equal}{*}

\begin{mlsysauthorlist}
\mlsysauthor{Jack Kosaian}{cmu,equal}
\mlsysauthor{Amar Phanishayee}{msr}
\end{mlsysauthorlist}

\mlsysaffiliation{cmu}{Carnegie Mellon University}
\mlsysaffiliation{msr}{Microsoft Research}

\mlsyskeywords{Machine Learning, MLSys}

\vskip 0.3in

\begin{abstract}
There has been significant progress in developing neural network architectures that both achieve high predictive performance and that also achieve high application-level inference throughput (e.g., frames per second). Another metric of increasing importance is GPU utilization during inference: the measurement of how well a deployed neural network uses the computational capabilities of the GPU on which it runs. Achieving high GPU utilization is critical to increasing application-level throughput and  ensuring a good return on investment for deploying GPUs.

This paper analyzes the GPU utilization of convolutional neural network (CNN) inference. We first survey the GPU utilization of CNNs to show that there is room to improve the GPU utilization of many of these CNNs. We then investigate the GPU utilization of networks within a neural architecture search (NAS) search space,  and explore how using GPU utilization as a metric could potentially be used to accelerate NAS itself. Our study makes the case that there is room to improve the inference-time GPU utilization of CNNs and that knowledge of GPU utilization has the potential to benefit even applications that do not target utilization itself. We hope that the results of this study will spur future innovation in designing GPU-efficient neural networks.
\end{abstract}
]

\printAffiliationsAndNotice{\mlsysInternship}
\section{Introduction} \label{sec:intro}
Convolutional neural networks (\cnns) have been widely used in many image- and video-processing tasks. In addition to designing \cnns that achieve high accuracy on a particular task, a large body of work has focused on designing \cnns that also operate with low latency or high application-level throughput (\eg frames per second)~\cite{redmon2016you,kang2017noscope}. The latency and throughput of \cnns has also been bolstered through advances in hardware, such as through GPUs that offer specialized compute units for accelerating neural networks (\eg NVIDIA's Tensor Cores)~\cite{nvidia-v100,nvidia-t4,nvidia-a100}.

An important, yet often overlooked, metric for \cnn inference on GPUs is \textit{\hardwareutilization}: the measure of the fraction of a GPU's peak computational performance that \cnn inference can sustain. Recent GPUs offer hundreds of \TFLOPS of compute bandwidth. It is critical that these GPUs be highly
utilized, with software running on an GPU ideally
achieving \TFLOPS near the GPU’s theoretical peak. Poorly utilizing a GPU leads to a poor return on investment for purchasing and deploying an accelerator, and leaves room on the table to further improve latency or throughput.

Despite the importance of \hardwareutilization in application performance and operational efficiency, \hardwareutilization is often overlooked in designing \cnns. While prior work has considered \hardwareutilization in highly specific settings~\cite{kosaian2021boosting}, the \hardwareutilization of general-purpose \cnns has  been less-widely studied. Other techniques aim to improve \hardwareutilization at the system level by efficiently co-executing multiple \cnns~\cite{jain2018dynamic,narayanan2018accelerating,wang2021horizontally}, but these works do not consider improving the \hardwareutilization of a single \cnn, which remains important in many settings.

This paper aims to fill this void by performing a multifaceted study of \hardwareutilization for \cnn inference. We focus specifically on GPUs, as they are widely used in datacenter settings for performing high-throughput inference. We decompose our study into three parts:

First, we survey the \hardwareutilization of \cnns used for image classification. We consider both \cnns developed manually as well as those found by architecture search techniques, and also consider \cnns that vary in terms of accuracy and throughput. Our survey reveals a mixed landscape of \hardwareutilization: some \cnns are capable of achieving high \hardwareutilization, while others fall significantly short, even when using large batch sizes. Notably, many of the \cnns that poorly utilize GPUs are on the Pareto frontier of other important metrics, such as accuracy and throughput. This leads one to question whether the accuracy and/or throughput of these \cnns could be further bolstered by increasing \hardwareutilization.

Based on this survey, we next investigate techniques to improve the \hardwareutilization of \cnns that poorly utilize GPUs. Specifically, we focus on \cnns that have low \ai---a property of a \cnn that drives its \hardwareutilization, and which will be defined in detail in \Section\ref{sec:background}. We investigate the potential for techniques previously proposed for increasing the \ai of small, specialized \cnns~\cite{kosaian2021boosting} to also provide benefit for general-purpose \cnns. We find that these techniques do indeed transfer to the domain of general-purpose \cnns in terms of improving \ai, throughput, and \hardwareutilization, though we do not consider whether the resultant \cnns maintain high accuracy. This offers promising potential for improving the \hardwareutilization of general-purpose \cnns.

Finally, we consider the role of \hardwareutilization in \nasFull (\nas). Alongside the analysis presented above, we additionally study the accuracy, throughput, and \hardwareutilization of \cnns a search space from \NATSbench~\cite{dong2021nats}. Our study reveals a positive correlation between \hardwareutilization and accuracy within this search space.  
This correlation leads us to consider whether \hardwareutilization can be used as a metric in place of accuracy during \nas to speed up the overall search process. \Hardwareutilization is a significantly cheaper metric to measure than accuracy during \nas; measuring \hardwareutilization requires only measuring throughput and analyzing the \FLOPs performed by a \cnn, whereas measuring accuracy requires training a candidate \nn. Thus, if possible, using \hardwareutilization in place of accuracy in \nas could reduce the search time in \nas. We investigate this question further by considering the use of \hardwareutilization as an approximate filtering metric to reduce the number of candidates in \nas required for full accuracy evaluation. While our preliminary results do not yield improvements in \nas time or search quality, we conclude that future \nas techniques should consider how they can leverage \hardwareutilization to potentially improve the \nas search process. 

Overall, this paper does not propose novel techniques with positive results. Rather, its aim is to survey \hardwareutilization in \cnn inference and to consider how it could be improved or used in other aspects of the machine learning lifecycle. This survey indicates that there is room for improvement in \hardwareutilization within \cnn inference. Improving \hardwareutilization offers promises not only for improved operational efficiency, but also for improving the throughput and/or accuracy of \cnns themselves. We hope that the our findings will spur further research in improving and exploiting \hardwareutilization.

\textbf{A note on the setting of this paper:} The research performed in this paper took place in the summer of 2021, as did the majority of the writing of this paper. We leveraged \cnns, GPUs, and software libraries that were widely used at the time, but they may either have been superseded or improved as of the release of this paper. Advancements in network design, hardware, and software may modify our findings.

\section{Background and Related Work} \label{sec:background}
In this section, we define \hardwareutilization, why it is important, and how to achieve high \hardwareutilization. We additionally describe related work on increasing \hardwareutilization.

\subsection{\Hardwareutilization} \label{sec:background:util}
Processors, whether CPUs, GPUs, or specialized accelerators, have a theoretical peak number of operations that they can perform per second. This metric is determined by the number of processing elements on the processor that can be used concurrently, the number of operations each processing element can perform per cycle, and the number of cycles the processor performs per second. This theoretical peak value is typically measured in terms of the number of floating-point operations that can be performed per second (\ie \FLOPS).

However, not all software running atop a processor is capable of achieving the theoretical peak \FLOPS of the processor. This gives rise to the definition of utilization we leverage in this work: the fraction of a processor's peak \FLOPS that a computation can sustain. As this paper focuses on GPUs, we will refer to this utilization as ``\hardwareutilization'' for the remainder of this discussion.

\textbf{Why is \hardwareutilization important?}
Maintaining high \hardwareutilization is important for multiple reasons. First, from the perspective of an application, underutilizing a GPU leaves application-level performance (\eg frames per second) on the table; underutilization means that one could potentially get more application-level value from the GPU on which they are already running. Second, underutilization leads to poor operational efficiency. GPUs are expensive and power hungry. Thus, underutilizing a GPU leads to a poor return on infrastructure investment, as well as less sustainable infrastructure. Thus, it is important for software to highly utilize GPUs.

\textbf{What is needed for a \cnn to highly utilize a GPU?}
A prerequisite for achieving the peak \FLOPS of a GPU is that the computation in question be compute bound: a compute-bound computation performs enough computation to keep all processing elements on a processor busy at all times. 

To determine whether a computation is compute bound, we turn to a popular performance model~\cite{williams2009roofline}. Intuitively, a compute-bound computational kernel is one that spends more time performing computation than it does reading/writing memory:
\begin{align*}
    \text{Compute time} & >  \text{Memory load/store time}
\end{align*}
\begin{align*}
    \frac{\text{\ops}}{\text{Compute Bandwidth}} & >  \frac{\text{Bytes}}{\text{Memory Bandwidth}}
\end{align*}
Here, ``\ops'' is the number of arithmetic operations performed by the kernel, ``Bytes'' is the amount of data the kernel transfers to/from memory, ``Compute Bandwidth'' is the GPU's peak \FLOPS, and ``Memory Bandwidth'' is the GPU's memory bandwidth (bytes/sec). Rearranging this inequality to pair properties of the kernel on the left-hand side and properties of the GPU on the right-hand gives: 
\begin{align}
     \frac{\text{\ops}}{\text{Bytes}} & >  \frac{\text{Compute Bandwidth}}{\text{Memory Bandwidth}}
    \label{equation:arithmetic_intensity}
\end{align}
The left-hand ratio of Equation~\ref{equation:arithmetic_intensity} is the kernel's \ai: the ratio between the \ops the kernel performs and the bytes it transfers to/from memory. The right-hand ratio is the GPU's \cmrFull (\cmr).

The inequality in Equation~\ref{equation:arithmetic_intensity} indicates that a computational kernel must have \ai higher than the \cmr of the GPU on which it executes in order to have the possibility of achieving the peak \FLOPS of the GPU. It is important to note that satisfying the inequality in Equation~\ref{equation:arithmetic_intensity} is only a prerequisite to achieving high utilization; doing so does not guarantee that one will achieve peak \FLOPS. A kernel satisfying this inequality can still poorly utilize a processor if the implementation of the kernel does not make efficient use of resources on the processor (\eg through not using vector instructions or inefficiently using the memory hierarchy).

\subsection{Effects of model and input size on \hardwareutilization} \label{sec:background:increase_util}
We now describe the effects that batch size and \cnn size have on \hardwareutilization.

In discussing each of these components, it is helpful to have a slightly deeper view of \ai for \cnns. The \ai of any layer of a \cnn can be (abstractly) written as:
\begin{equation}
\label{equation:conv_abstract}
     \frac{\text{\ops}}{\text{Input bytes} + \text{Weight bytes} + \text{Output bytes}}
\end{equation}
``\ops'' is, again, the number of  arithmetic operations performed by the layer. ``Input bytes'' is the total size of the layer's input activations, ``Output bytes'' is the total size of output activations written by the layer to memory for processing by the next layer, and ``Weight bytes'' is the total size of the layer's weights.

In defining the aggregate \ai of a \cnn as a whole, one sums the \ops performed across all layers, sums the bytes transferred to/from memory by all layers, and divides these components. This metric is not as useful as the \ais of individual layers of a \cnn, as \cnn inference is performed on a per-layer basis. Nevertheless, it gives a broad notion of how compute- or memory-bandwidth-bound a \cnn as a whole is likely to be. Finally, we note that our analysis here assumes that common optimizations that increase \ai are performed, such as fusing non-linear layers to the preceding linear layers to minimize memory traffic.

\paragraph{Effect of batch size.} The input to a \cnn is a batch of $\symBatchSize$ images each with resolution $\symHeight \times \symWidth$ and $\symChannelsIn$ channels (\eg $\symChannelsIn = 3$ for RGB images). Here, we will focus on the effect of batch size $\symBatchSize$ on the \ai of a \cnn, as high-throughput vision applications often leverage large batch sizes.

A layer in a \cnn operates (abstractly) by (1)~loading the layer's weights from memory, (2)~loading the input activations to the layer from memory, (3)~computing over the input activations using layer weights, and (4)~writing the outputs of the layer to memory.

Each of steps 2, 3, and 4 above scale linearly with increasing batch size, while step~1 remains a constant cost regardless of the batch size used. Translating these scaling factors to Equation~\ref{equation:conv_abstract}, we see that increasing batch size leads to a linear increase in the numerator and a sub-linear increase in the denominator (because ``Weight bytes'' does not increase with batch size). Thus increasing batch size increases \ai. The increase in \ai from increased batch size is intuitive: operating at a larger batch size better amortizes the cost of loading a layer's weights from memory.

However, increasing batch size eventually leads to diminishing returns in \ai: once batch size has been increased to the point in which loading layer weights accounts for negligible memory traffic, further increasing batch size provides an insignificant increase in \ai. Prior work has illustrated this limit for \cnns developed through model specialization~\cite{kosaian2021boosting}. For these \cnns, \ai still remains lower than the \cmr of server-grade GPUs even when operating at large batch sizes.

\paragraph{Effect of \cnn size.}
Large \cnns, such as those with more channels per convolutional layer, typically perform more \ops than small \cnns. As a result, they typically have higher \ai, and thus also typically better utilize GPUs. However, this increased \op count typically comes at the expense of lower application-level throughput or higher latency.

On the other hand, decreasing the size of a \cnn typically reduces the number of \ops it performs. This often leads to higher application-level throughput or lower latency, at the expense of lower \ai, and thus lower \hardwareutilization. For example, techniques like model scaling (\eg through EfficientNets~\cite{tan2019efficientnet}) and model specialization produce \cnns that can operate with higher throughput than larger \cnns, but which typically poorly utilize GPUs~\cite{kosaian2021boosting}.

\subsection{Related work on high \hardwareutilization} \label{sec:background:related}
We now highlight related work aimed at improving the \hardwareutilization of \cnns.

\paragraph{Improving utilization via multi tenancy.}
One approach to increase \hardwareutilization is to concurrently execute multiple \nnsFull on a single GPU at once. This technique has been explored both for training \nnsFull, such as through scheduling systems~\cite{xiao2018gandiva,gu2019tiresias}, as well as for inference, by fusing and co-executing similar layers of distinct \nnsFull together~\cite{narayanan2018accelerating,jain2018dynamic,wang2021horizontally} or through better management of GPU resources~\cite{yu2020salus}.

In contrast to these works, our focus in this work is on analyzing the \hardwareutilization of performing inference over a single \cnn, rather than a group of co-scheduled \cnns. Techniques used to improve the \hardwareutilization of a single \cnn can be used alongside these co-scheduling techniques.

\paragraph{Designing hardware-efficient \cnns.}
An alternative line of work has focused on designing \cnns that efficiently make use of a given device. This has been leveraged to develop low-latency \cnns for mobile deployments (\eg~\cite{wu2019fbnet}), as well as architecture search techniques that can achieve high performance across a variety of hardware backends~\cite{cai2019once}.

Along this same line of work, multiple works have considered developing GPU-efficient \cnns. These works either optimize for metrics that are proxies for \hardwareutilization (\eg frames/sec. on a GPU)~\cite{molchanov2021hant,ridnik2021tresnet}, or by optimizing for a metric closely related to \hardwareutilization, such as \ai~\cite{zhou2018resource}. Other work has proposed transformations to \cnns in specific settings to improve their GPU utilization~\cite{kosaian2021boosting}.

In contrast to these works, our focus in this paper is primarily in surveying the \hardwareutilization of \cnn inference. In doing so, we evaluate some of the networks discovered by techniques listed above.

\paragraph{System support for GPU-efficient inference.}
There have been many recent works that aim to make better use of the underlying hardware on which \nnFull inference is executed~\cite{chen2018tvm,nvidia-tensorrt}. However, as described above, achieving high \hardwareutilization requires not only that software be optimized, but also that a \cnn have a high enough \ai. Thus, in this work, we focus on the \hardwareutilization of \cnn inference primarily by studying the \ai of \cnns. However, we note that all of our results involve executing \cnns when using the GPU-optimized TensorRT SDK~\cite{nvidia-tensorrt}.
\section{Analyzing \hardwareutilization in \cnn inference} \label{sec:survey}
We begin our study of \hardwareutilization in \cnn inference by surveying the \hardwareutilization of a variety of \cnns. Through this study, we establish that these \cnns operate within a wide spectrum of \hardwareutilization. We conclude by discussing and evaluating a potential opportunity to improve \hardwareutilization.

As described in \Section\ref{sec:intro}, the research performed in this paper took place in the summer of 2021. We leveraged \cnns, GPUs, and software libraries that were widely used at the time, but they may either have been superseded or improved as of the release of this paper; e.g., our \cnns do not reflect the recent trend of using transformers in vision tasks. Advancements in network design, hardware, and software may modify our findings.

\subsection{Setup} \label{sec:survey:setup}
We analyze the \hardwareutilization of \cnns by executing them atop a V100 GPU through the TensorRT SDK and when using FP16 datatypes, which involves the use of Tensor Cores~\cite{nvidia-tensorcores}. We choose this setting because, at the time when this research was performed, the V100 was a widely-deployed GPU in datacenters today used for high-throughput applications. Our microbenchmarks reveal a maximum achievable performance of 100 \TFLOPS on the device used for evaluation, which is on par with that reported in prior work~\cite{jia2018dissecting}. We execute each \cnn at the maximum batch size that it can fit within the GPUs memory. Operating at large batch size is common for high-throughput applications.

We analyze 103 \cnns used for image classification from the \timm repository~\cite{wightman2019timm}, as well as those used for the V100 in the once-for-all network~\cite{cai2019once}. The \timm repository contains implementations of a large collection of \cnns, many of which have, at some point in time, achieved state-of-the-art accuracy on popular tasks, or state-of-the-art results on the \pf of accuracy and latency/FLOP count. \cnns within this repository include both those that have been hand crafted as well as those that have been discovered through architecture search techniques. We only include \cnns that compiled with TensorRT at the time of our experimentation. We additionally measure the performance of \cnns specialized fro the V100 from  Once-For-All networks~\cite{cai2019once}. A list of all \cnns considered is provided in \Section\ref{app:survey}.

\paragraph{Measuring \hardwareutilization.} The metric of interest for \hardwareutilization is the fraction of the peak achievable \FLOPS of the V100 GPU that a \cnn can sustain. To measure the \FLOPS achieved by the \cnn, we measure the throughput of the \cnn in terms of images/second over 10000 batches and multiply this by the \FLOP count of the \cnn when operating over a single image. This value is then divided by the GPU's peak achievable \FLOPS to obtain \hardwareutilization as a fraction. We find the V100 to achieve 100 \TFLOPS in FP16 when utilizing Tensor Cores on a large matrix multiplication, and use this as the peak achievable \FLOPS of the device. We use input and output sizes commonly used for ImageNet~\cite{russakovsky2015imagenet}: images are of resolution $224 \times 224$, and each \cnn produces a prediction vector over 1000 classes.

\paragraph{Reporting accuracy.} We additionally report the accuracy of \cnns on the ImageNet dataset. We do not train these \cnns ourselves, but rather report the accuracies of these \cnns that have been attained within the \timm and once-for-all network repositories.

\begin{figure*}[!t]
	\centering
    \begin{subfigure}{0.3\textwidth}
        \includegraphics[width=\textwidth]{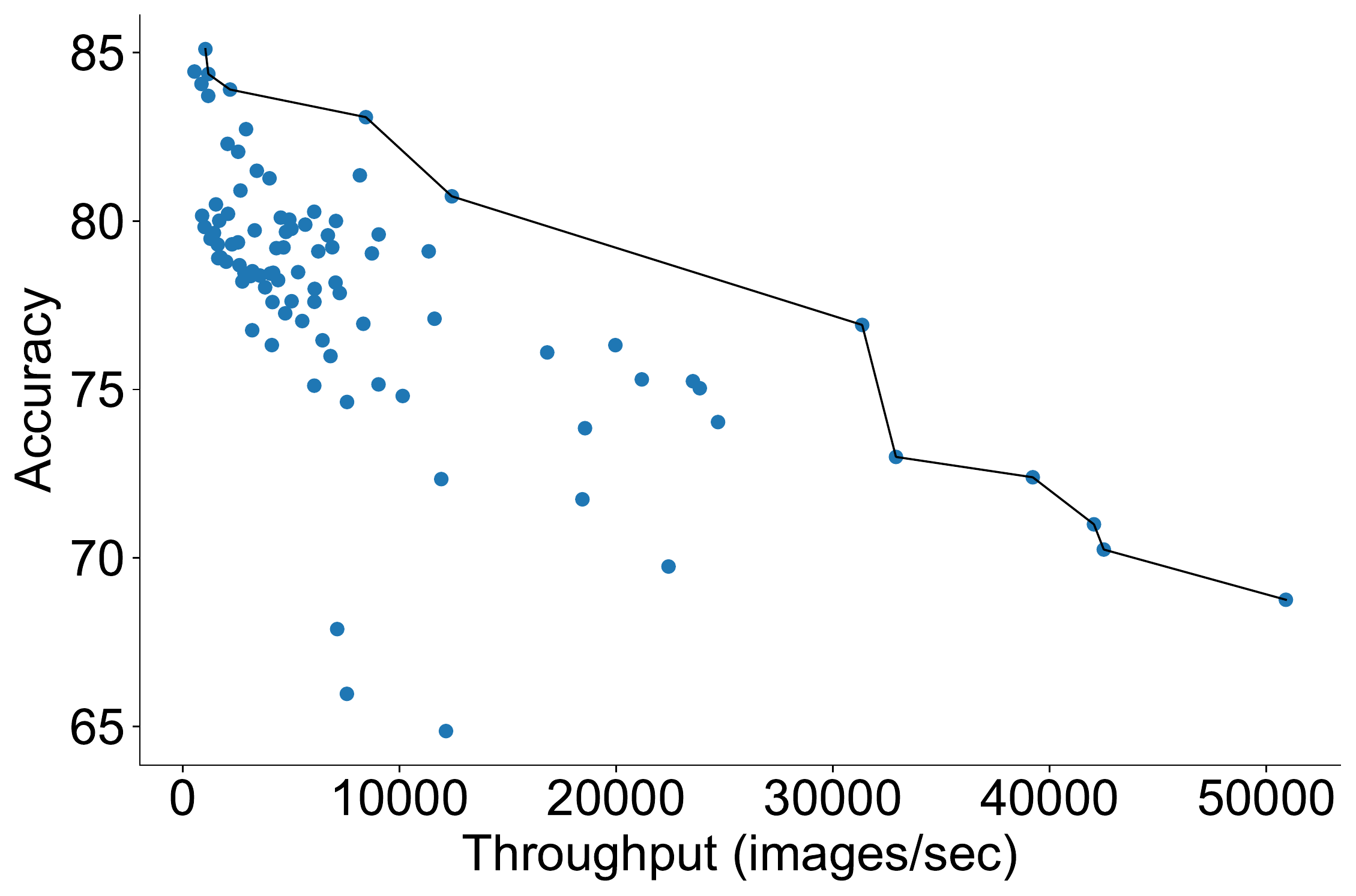}
    	\caption{Throughput-accuracy \pf}
    	\label{fig:survey:frontiers:tput_acc}
    \end{subfigure}
    \hfill
    \begin{subfigure}{0.3\textwidth}
        \includegraphics[width=\textwidth]{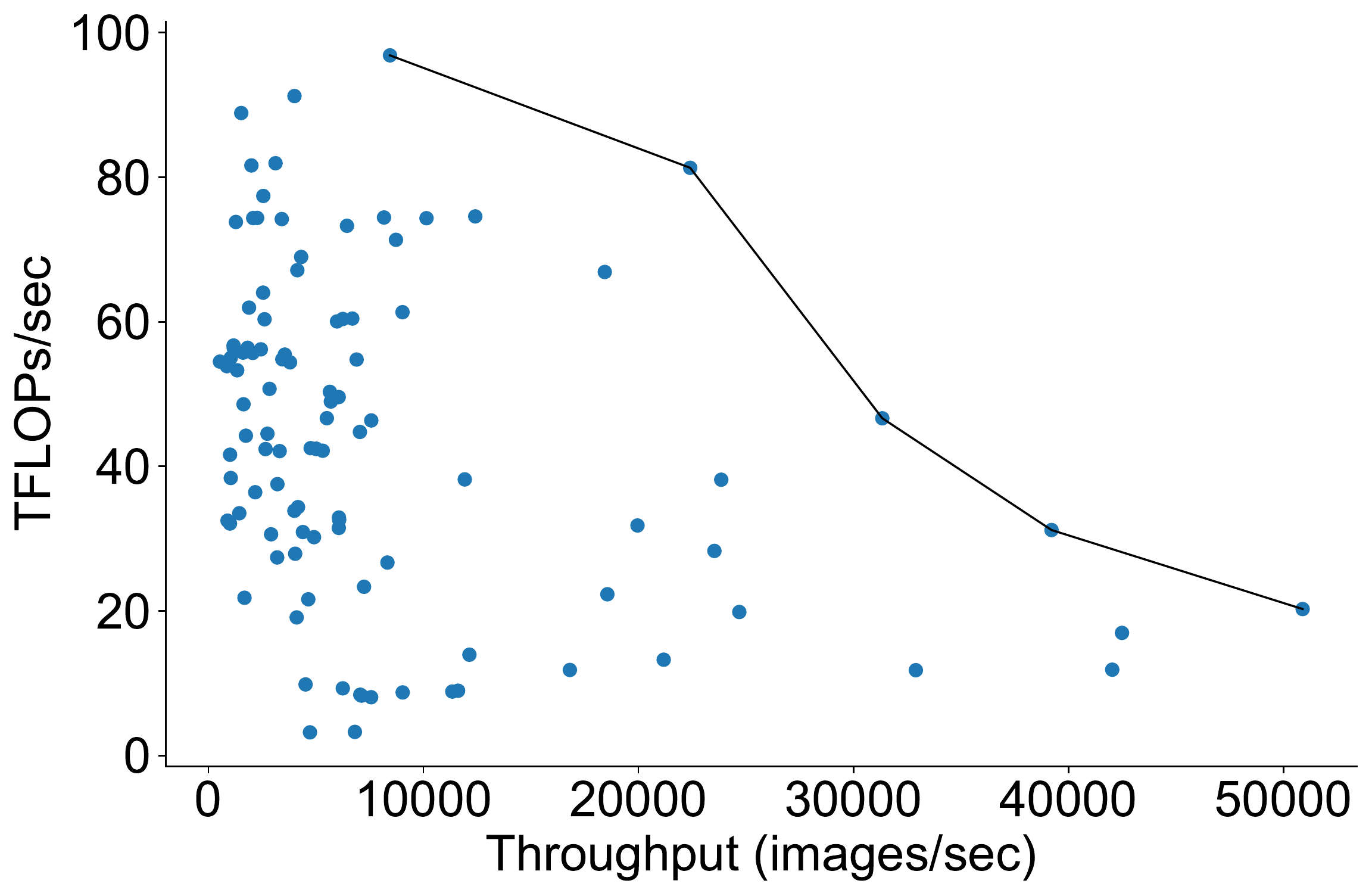}
    	\caption{Throughput-utilization \pf}
    	\label{fig:survey:frontiers:tput_tflops}
    \end{subfigure}
    \hfill
    \begin{subfigure}{0.3\textwidth}
        \includegraphics[width=\textwidth]{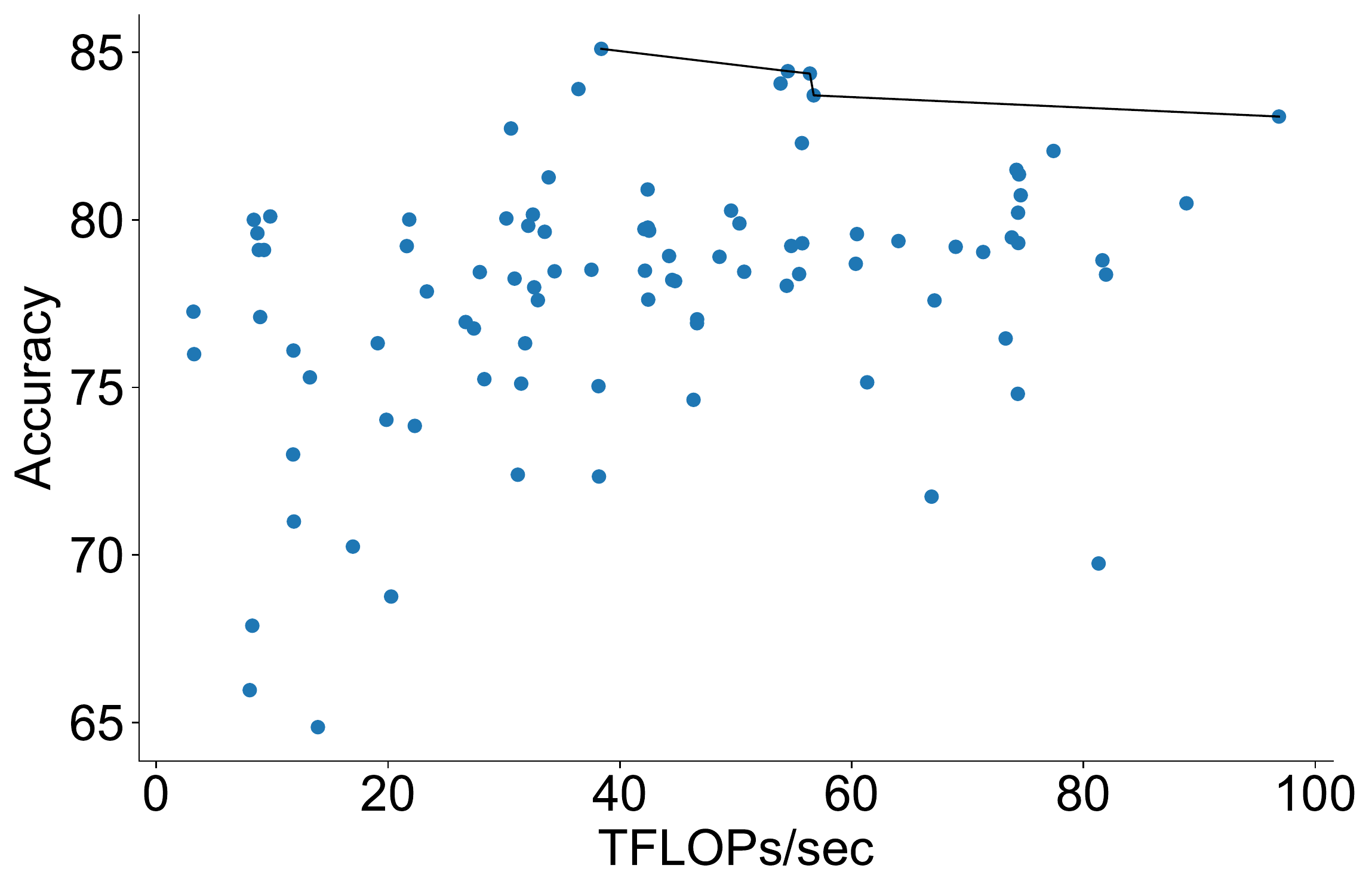}
    	\caption{Utilization-accuracy \pf}
    	\label{fig:survey:frontiers:tflops_acc}
    \end{subfigure}
    \caption{\pfs between metrics of accuracy, throughput, and \hardwareutilization. \cnns on the \pf are connected by a solid black line.}
    \label{fig:survey:frontiers}
\end{figure*}

\subsection{Analysis of \cnn \hardwareutilization}
\Figure\ref{fig:survey:frontiers} plots the accuracy, throughput, and \hardwareutilization of all \cnns considered. Plotting these \cnns along any two of these three metrics results in a two-dimensional \pf. Overall, we find that only one \cnn, TResNet-M, sits along all three \pfs. It is interesting to note that the TResNet family of \cnns was designed specifically to operate efficiently on GPUs~\cite{ridnik2021tresnet}.

We next analyze each of the two-dimensional \pfs displayed in \Figure\ref{fig:survey:frontiers}.

\paragraph{Throughput-accuracy.} \Figure\ref{fig:survey:frontiers:tput_acc} plots each \cnn according to its achieved throughput and accuracy. This is a popular frontier to consider, as throughput is an important application-level metric targeted by many systems. We observe a general trend that \cnns with higher throughput typically have lower accuracy. This is likely due to these high-throughput \cnns being smaller in size, both in terms of parameters and \FLOP count. Reducing \cnn size often brings improvements in throughput or latency at the expense of lower accuracy. We additionally observe a large cluster of \cnns that have lower throughput and higher accuracy. This is likely an artifact of sampling bias, as many of the \cnns we considered were designed primarily to achieve high accuracy, but did not necessarily consider throughput as a metric of importance.

\paragraph{Throughtput-utilization.}
\Figure\ref{fig:survey:frontiers:tput_tflops} plots each \cnn according to its achieved throughput and \hardwareutilization. This plot looks similar to that in \Figure\ref{fig:survey:frontiers:tput_acc}, but with \cnns having a more diverse range of \hardwareutilizations, plotted on the y-axis. We make two primary observations from this figure: 

(1)~The \cnns surveyed here tend not to have both high throughput and high \hardwareutilization. Rather, the \cnns that achieve higher throughput tend to have lower \hardwareutilization. This is likely due to these \cnns performing fewer \FLOPs than larger \cnns, which can increase throughput, but often at the expense of \hardwareutilization. 

(2)~Sampled \cnns that are on the lower end of the spectrum in terms of throughput have a wide variety of \hardwareutilizations. There are many \cnns that achieve throughput of less than 10000 samples per second. Among these, we see \cnns that vary in \hardwareutilization from near 0\% to near 90\%. Given that the \cnns in this range have similar throughput, the primary factor leading to their varying \hardwareutilizations must be their \FLOP count. This indicates either that (a)~those \cnns within this band that achieve higher \hardwareutilization leverage operations that are more efficient on the GPU, or (b)~those \cnns within this band that achieve lower \hardwareutilization leverage operations that are less efficient on the GPU. It is likely a case that we see a mix of these two options for different sampled \cnns.

\paragraph{\hardwareutilization-accuracy.} \Figure\ref{fig:survey:frontiers:tflops_acc} plots each \cnn according to its achieved \hardwareutilization and accuracy. We generally see a positive correlation between \hardwareutilization and accuracy. We investigate and exploit this correlation further in \Section\ref{sec:accel}.

\begin{figure*}[!t]
	\centering
    \begin{subfigure}{0.4\textwidth}
        \includegraphics[width=\textwidth]{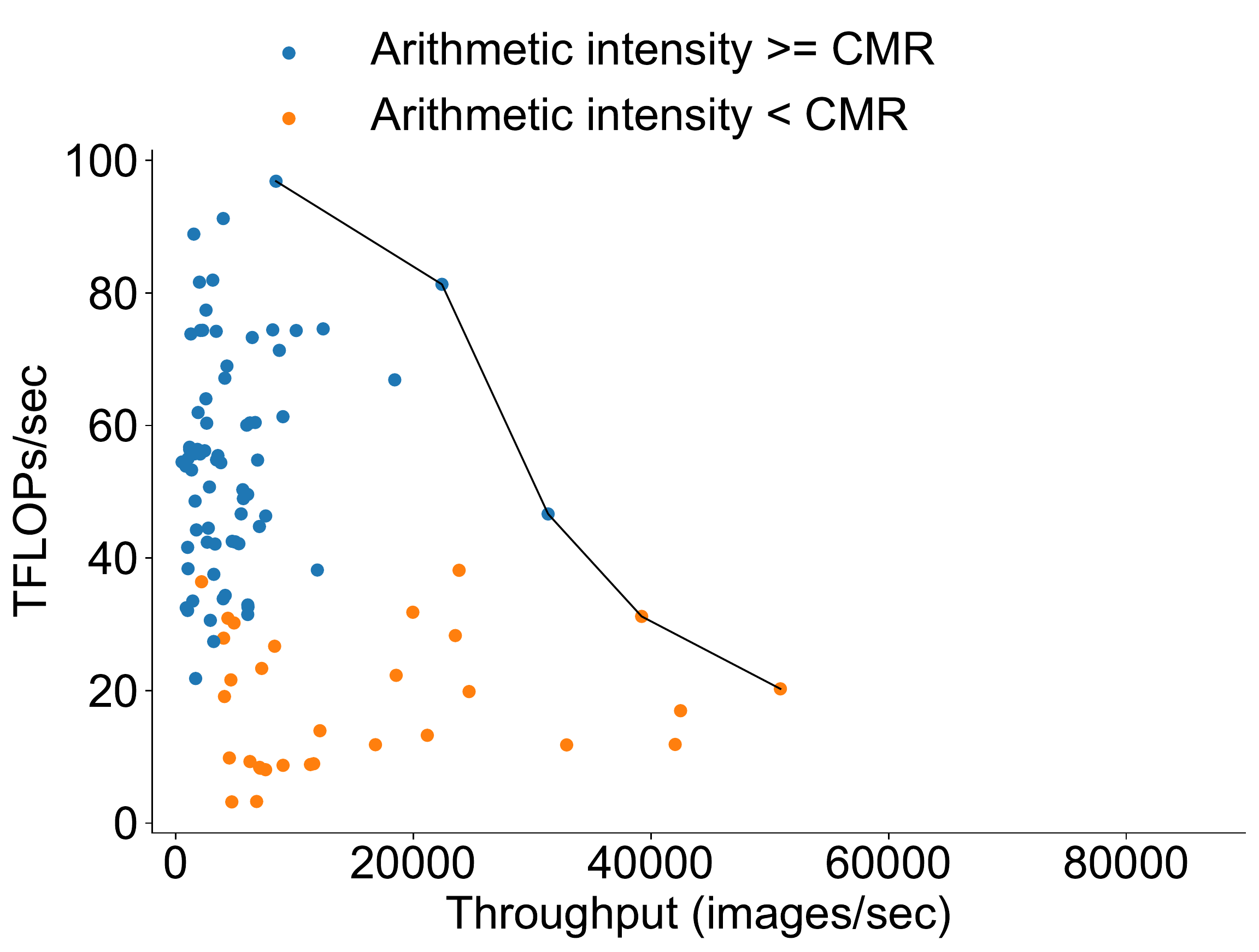}
    	\caption{Throughput-\hardwareutilization \pf with \cnns grouped by \ai.}
    	\label{fig:survey:ai:tput_tflops}
    \end{subfigure}
    \hfill
    \begin{subfigure}{0.4\textwidth}
        \includegraphics[width=\textwidth]{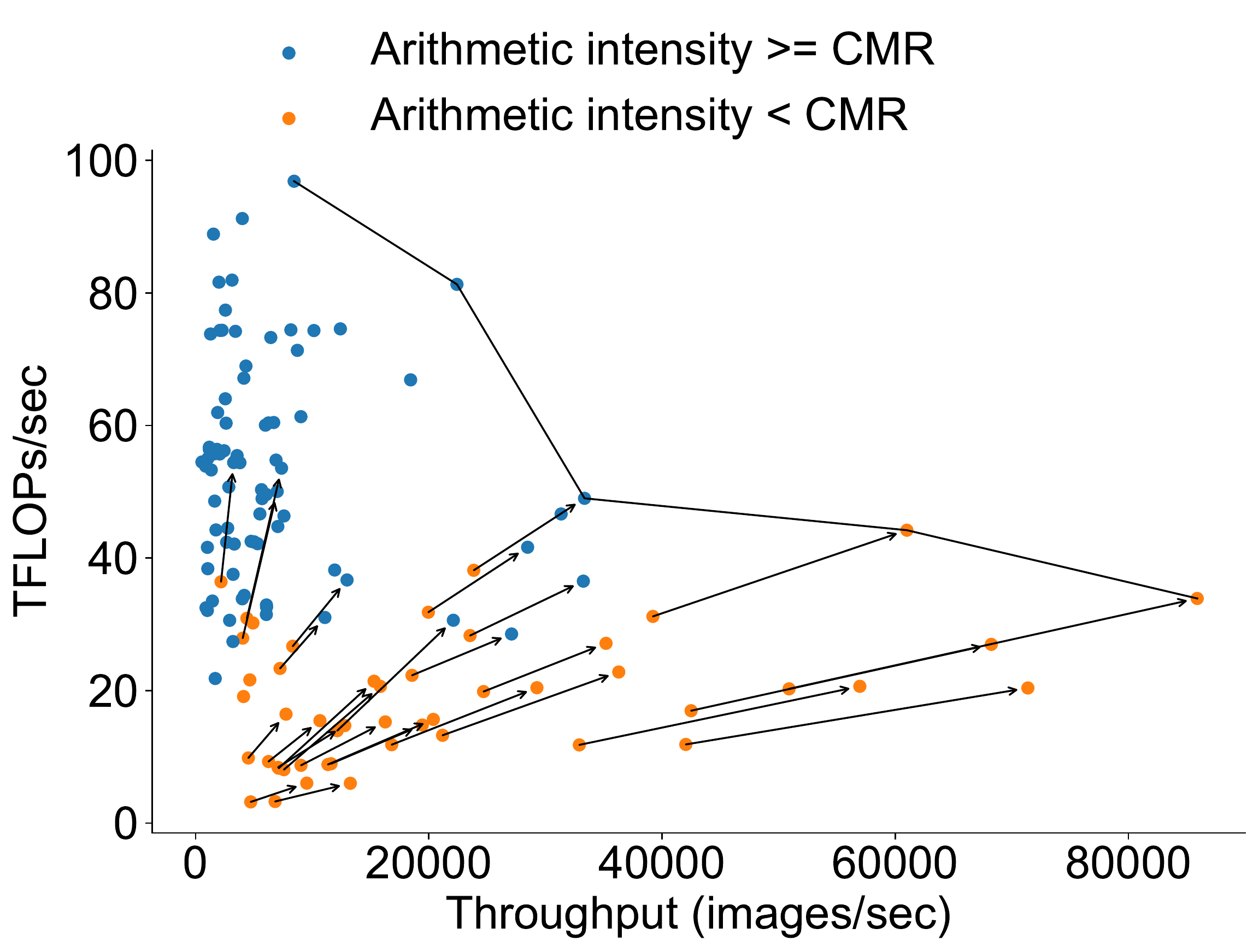}
    	\caption{Throughput-\hardwareutilization \pf with \cnns grouped by \ai after performing folding with parameter $f = 4$ for \cnns with \ai below the \cmr of the V100.}
    	\label{fig:survey:ai:tput_tflops_fold}
    \end{subfigure}
    \caption{\pfs between throughput and \hardwareutilization when grouping \cnns based on whether their FP16 \ai is greater or less than the the FP16 \cmr of the V100 GPU (139). Figure (b) shows the same plot with the changes in throughput and \hardwareutilization made possible for performing a folding transformation with $f = 4$ on those \cnns with low \ai.}
    \label{fig:survey:ai}
\end{figure*}
\subsection{Opportunities to increase \hardwareutilization} \label{sec:survey:increasing_util}
Recall the prerequisite outlined in \Section\ref{sec:background:util} for achieving high \hardwareutilization: \ai must be greater than the \cmrFull (\cmr) of the GPU on which a \cnn runs. We next analyze the \ais of the \cnns surveyed above to determine opportunities to improve \hardwareutilization.

\Figure\ref{fig:survey:ai:tput_tflops} plots the \pf between throughput and accuracy (\ie \Figure\ref{fig:survey:frontiers:tput_tflops}) but with each \cnn marked by whether its FP16 \ai is greater or less than the FP16 \cmr of the V100 GPU (139). As expected, we find that those \cnns that are on the higher end of the spectrum in terms of \hardwareutilization typically have \ai greater than the \cmr of the V100, while those with lower \hardwareutilization typically have \ai below the \cmr of the V100.

The many \cnns with \ai lower than the \cmr of the V100 offer opportunities for potentially improving both throughput and \hardwareutilization by increasing \ai. We next evaluate how a recently-proposed technique, FoldedCNNs~\cite{kosaian2021boosting}, could potentially be used to transform these \cnns to increase \ai. FoldedCNNs involve restructuring the inputs to a \cnn as well as small changes to the \cnn itself. Under a FoldedCNN, rather than operating over a batch of $N$ images each with $C$ channels (\eg $C=3$ for RGB images), a FoldedCNN instead operates over a batch of $\frac{N}{f}$ inputs, each with $Cf$ channels, consisting of $f$ images stacked along the channels dimension. The FoldedCNN then infers over these ``combined'' images. In addition to this restructuring of inputs, a FoldedCNN also increases the width of each layer of the \cnn by a factor of $\sqrt{f}$ (\eg increasing the number of input and output channels of a convolutional layer by a factor of $\sqrt{f}$). This overall transformation performed by FoldedCNNs is referred to as ``folding.'' Under certain settings, this transformation is proven to transform a \cnn such that it performs nearly the same number of \FLOPs, but with a reduction in memory traffic by a factor of $\sqrt{f}$. This leads to FoldedCNNs increasing \ai under these settings by a factor of $\sqrt{f}$.

These modifications made by a FoldedCNN require the \cnn to be retrained, and the new structure of inputs of a FoldedCNN appears to make the task of a FoldedCNN more challenging, which often leads to lower accuracy. For the purposes of our discussion here, we will focus only on the potential of FoldedCNNs to improve the \hardwareutilization of \cnns surveyed above, leaving accuracy considerations separate.

\Figure\ref{fig:survey:ai:tput_tflops_fold} shows the throughput and \hardwareutilization achieved by each \cnn when those \cnns with \ai below the CMR of the V100 have been folded using parameter $f = 4$. For each \cnn that has been modified, we connect an arrow starting from the \cnn's original throughput and \hardwareutilization to the new throughput and \hardwareutilization achieved after folding. We make the following observations:
\begin{denseenum}
    \item \textit{Techniques to increase \ai expand the \pf between throughput and \hardwareutilization.} \Figure\ref{fig:survey:ai:tput_tflops_fold} shows that the application of folding significantly increases the throughput and \hardwareutilization of \cnns on the \pf. For example, one \cnn on the \pf increases in both throughput and \hardwareutilization by a factor of roughly 1.5.
    
    \item \textit{Further room for improving \hardwareutilization by increasing \ai remains.} While many of the \cnns transformed in \Figure\ref{fig:survey:ai:tput_tflops_fold} significantly increase in throughput and \hardwareutilization, many of the transformed \cnns still have \ai lower than the CMR of the V100 (\ie the dots at the beginning and end of the arrow have the same color). This indicates that further improvements in \ai may lead to further increases in throughput and \hardwareutilization, if done judiciously. One technique to do so could be to increase the parameter $f$ in folding, as the theoretical increase in \ai resulting from folding scales with $\sqrt{f}$. However, doing so is likely to cause significant accuracy degradation; in the original FoldedCNNs evaluation, accuracy began to noticeably degrade at $f = 4$~\cite{kosaian2021boosting}. If this challenge of maintaining high accuracy could be circumvented, further increasing \ai via folding would be a natural solution.
\end{denseenum}
\section{Accelerating \nas by using \hardwareutilization?} \label{sec:accel}
We now switch gears to analyzing another finding from \Section\ref{sec:survey}: the positive correlation between accuracy and \hardwareutilization observed in \Figure\ref{fig:survey:frontiers:tflops_acc}. Specifically, we explore the opportunity to potentially use this correlation to leverage \hardwareutilization as a lightweight replacement metric for accuracy in traditional \nas search.

\begin{figure}[!t]
	\centering
    \includegraphics[width=0.8\linewidth]{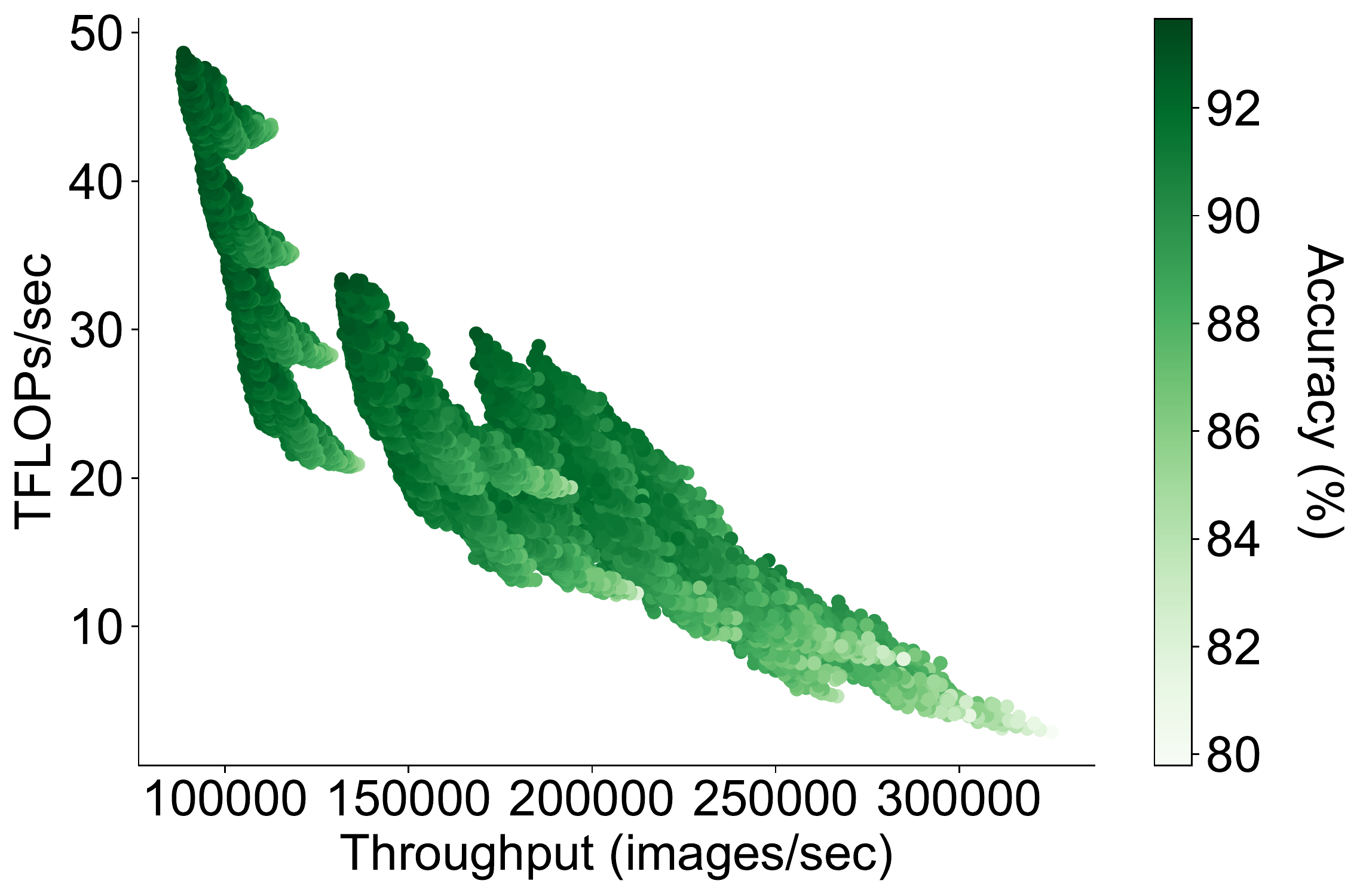}
    \caption{Throughput, \hardwareutilization, and accuracy of each \cnn from the \NATSbench size search space. Note that each \cnn achieves less than 60\% of the peak \FLOPS of the V100 GPU.}
    \label{fig:nas:survey}
\end{figure}

\subsection{Positive correlation between accuracy and \hardwareutilization} \label{sec:accel:corr} \label{sec:nas:survey}
\Figure\ref{fig:survey:frontiers:tflops_acc} illustrated a positive correlation between the \hardwareutilization of a \cnn and the accuracy achieved by the \cnn. 

To further analyze this correlation in the context of \nas, we analyze \cnns from the \NATSbench~\cite{dong2021nats} ``size search space.'' \NATSbench is a \nas benchmarking suite that contains the accuracies and model statistics (\eg parameter count) for a large number of \cnn architectures sampled from a pre-specified search space. This allows researchers to experiment with changes in the \nas search algorithm itself quickly without the need to train each candidate architecture.

We specifically focus on the \NATSbench ``size search space.'' Each \cnn within this search space has the same depth (in terms of number of convolutional layers) and the same layer types (\eg filter spatial resolution). However, each architecture differs in the number of channels per layer. We leverage the accuracy measurements reported within \NATSbench for the CIFAR-10 dataset. We measure throughput and \hardwareutilization using the same evaluation setup described in \Section\ref{sec:survey:setup}.

\Figure\ref{fig:nas:survey} plots the throughput, \hardwareutilization, and accuracy of each \cnn from the \NATSbench size search space. While there are a number of trends of interest in this plot, we focus on one finding: \textit{as \hardwareutilization increases, accuracy typically increases.} This is illustrated in \Figure\ref{fig:nas:survey} by noting that the shade of each point in the plot typically gets darker with increasing \hardwareutilization. This observation is likely explained by an increase in \FLOP count for \cnns with higher \hardwareutilization, as increasing \FLOP count is likely to increase both accuracy and \hardwareutilization. Furthermore, within the vertical bands of \cnns with near equal throughput in \Figure\ref{fig:nas:survey}, we typically observe an increase in accuracy with increasing \hardwareutilization.

\textbf{Takeaway.} Both across general-purpose \cnns (\Section\ref{sec:survey}) as well as a benchmark \nas search space, we find a positive correlation between \hardwareutilization and accuracy. In the remaining portions of this section, we investigate how this correlation can potentially be exploited to reduce the number of samples in \nas that require accuracy evaluation.

\subsection{Background on sample-based \nas} \label{sec:accel:background}
We first provide background on the sample-based \nas procedure that we aim to accelerate and the target \pf that we aim to optimize for.

\subsubsection{Sample-based \nas objective and procedure} \label{sec:accel:background:objective}
We consider \nas to be parameterized by a \searchspace $\symSearchSpace$ consisting of \nnFull architectures to be considered, along with a set of evaluation functions $\symEvalSet$ that are used to score candidate architectures from the \searchspace. The goal of \nas is to find the set of \nnsFull from $\symSearchSpace$ that lie on the \pf of $\symEvalSet$. 
Concretely, the \pf contains all \nnsFull which are not dominated by another \nnFull in all evaluation functions in $\symEvalSet$.

We now describe the search procedure used in sample-based \nas. For a comprehensive survey of other \nas methods, please see \citet{elsken2019neural}. To find the desired \pf,  sample-based \nas iteratively selects a \nnFull $\symSample$ from $\symSearchSpace$, evaluates $\symSample$ on each evaluation function in $\symEvalSet$, adds $\symSample$ to its running \pf, and (optionally) updates its selection criteria based on the evaluation of $\symSample$. There are many methods that can be used to select \nnsFull from the \searchspace, such as random search~\cite{li2020random}, evolutionary search~\cite{real2019regularized}, and reinforcement learning~\cite{zoph2016neural}. In this work, we abstract away the exact technique being used to guide the search process, as our focus is on the evaluation functions themselves.

Associated with each evaluation function $\symEvalI \in \symEvalSet$ is a time that it takes to perform the evaluation $\symEvalTimeI$.\footnote{Here, we make a simplifying assumption that each evaluation function takes a constant amount of time, regardless of the \nnFull over which it operates. This will not be the case in practice, as evaluation functions such as accuracy will take variable amounts of time, depending on the \nnFull being evaluated. For the purposes of the present discussion, which will be about comparing evaluation function times across evaluation functions, such a simplifying assumption suffices.} The time that it takes to perform an iteration of the above search procedure is determined by either the sum of all such evaluation times (if evaluation functions are executed serially), or the maximum evaluation time (if evaluation functions are executed in parallel). To bound the overall time that \nas can run, \nas is typically parameterized with a ``time budget'' $\symTimeBudget$ after which the search is terminated. Thus, the \pf returned by \nas may not be the true \pf of the entire \searchspace $\symSearchSpace$ in cases where \nas is unable to exhaustively enumerate through the \searchspace within the prescribed time budget.

\subsubsection{Accuracy, \throughput frontier} \label{sec:accel:background:frontier}
In this work, we focus on one concrete instantiation of the sample-based \nas procedure described in \Section\ref{sec:accel:background:objective}: that when the target \pf is of accuracy and \throughput. In this scenario, $\symEvalSet$ consists of two evaluation functions: one that evaluates the accuracy of a given \nnFull (denoted $\symEvalFnAcc$), and one that evaluates the inference-time \throughput of the \nnFull in inputs/sec (denoted $\symEvalFnTput$). A \nas search procedure may define an objective ranking function for a given \nnFull based on a combination of these two evaluation functions. For example, a ranking function $\symRank$ derived from that used by \citet{tan2019mnasnet} combining these evaluation functions might be:
\begin{equation}
    \symRank(\symSample) = \frac{\symEvalFnAcc(\symSample)}{100}\times\left[\frac{\symEvalFnTput(\symSample)}{\symTputGoal}\right]^\symExpTput
    \label{eqn:nas_eval}
\end{equation}
where $\symTputGoal$ is a constant, target \throughput, and $\symExpTput$ controls how much weight to give \throughput in ranking. \Algorithm\ref{alg:sample_og} illustrates this overall search procedure.

\begin{algorithm}[t]
\caption{Sample-based \nas for finding the accuracy-\throughput \pf}\label{alg:sample_og}
\begin{algorithmic}
\STATE $\symEvalTime \gets 0$
\WHILE{$\symEvalTime < \symTimeBudget$}
    \STATE $\symSample \gets sample(\symSearchSpace)$
    \STATE $\symEvalResultAcc \gets \symEvalFnAcc(\symSample)$
    \STATE $\symEvalResultTput \gets \symEvalFnTput(\symSample)$
    \STATE $\symRank_{\symSample} \gets \symRank(\symEvalResultAcc, \symEvalResultTput)$
    \STATE $\symEvalTime \gets \symEvalTime + max(\symEvalTimeAcc, \symEvalTimeTput)$ \COMMENT{Assumes $\symEvalFnAcc$ and $\symEvalFnTput$ run in parallel}
    \IF{$onFrontier(\symEvalResultAcc, \symEvalResultTput)$}
        \STATE $addToFrontier(\symSample, \symEvalResultAcc, \symEvalResultTput)$ \COMMENT{Removes points from running frontier, if necessary}
    \ENDIF
    \STATE \COMMENT{Optionally update sampling function based on $\symRank_{\symSample}$ (\eg if using reinforcement-learning-based search)}
\ENDWHILE
\end{algorithmic}
\end{algorithm}

\textbf{Dominant cost of sample-based \nas.} 
Ideally, sample-based \nas would be able to sample all possible \nnsFull from the \searchspace. However, this is precluded by the cost of evaluation functions, the large sizes of \searchspaces used in practice, and limitations on the compute resources that can be devoted to \nas. 

In particular, the primary bottleneck that contributes to the large time and resource cost of sample-based \nas is \textit{accuracy evaluation} (\ie $\symEvalFnAcc$). Evaluating the accuracy of a sample requires training the sampled \nnFull, which can take on the order of GPU-days, if done to convergence. 
In contrast, evaluating inference-time \throughput (\ie $\symEvalFnTput$) is much less expensive, taking at most minutes on a GPU.

\textbf{Reducing costs.}
Based on the dominance of evaluating accuracy on the overall time and resource cost of sample-based \nas, it is clear that the time and resource cost of sample-based \nas could be greatly reduced by either (or both of): (1)~reducing the time and resources required for accuracy evaluation (\ie reducing $\symEvalTimeAcc$), (2)~reducing the number of samples that require accuracy evaluation. These two opportunities toward speeding up sample-based \nas complement one another.

A number of works have developed techniques toward opportunity~(1) of reducing the time and resource costs of accuracy evaluation. The key approach used therein is to approximate accuracy evaluation. Examples of such techniques include training on a cheaper, proxy dataset (\eg CIFAR-10 instead of ImageNet) (\eg~\cite{liu2018darts}); training with a smaller \nnFull than the true \nnFull sampled from the \searchspace; training on smaller inputs than those used in the true task (\eg reducing image resolution); stopping training earlier than reaching final convergence (\eg~\cite{li2020random}); and sharing weights between distinct sampled \nnsFull (\eg~\cite{pham2018efficient}). These techniques have some potential downsides, such as the requirement for the availability of a representative proxy dataset (which may not be available in all domains), and assumptions about the fidelity of the  accuracy approximation being made. However, as a whole, these techniques remain promising for speeding up sample-based \nas.

Toward opportunity~(2), reducing the number of \nnsFull for which accuracy evaluation is required, the primary techniques used involve improving the sampling algorithm itself. This may entail using some form of learning in the sampling procedure in attempt to determine which samples are likely to achieve high performance along the \pf, based on samples that one has already evaluated.

In this section, we explore an alternative path toward opportunity~(2) that does not require expensive learning procedures. We next describe this approach.

\subsection{Opportunity: leveraging approximate filtering} \label{sec:accel:opportunity}
We explore the use of \textit{approximate filtering} to accelerate sample-based \nas. In this subsection, we provide background on approximate filtering, how it could be applied within sample-based \nas, and the potential reduction in \nas time and resource cost it could introduce.

\subsubsection{Approximate filtering} \label{sec:accel:opportunity:approximate_filtering}
Approximate filtering is a common technique leveraged in data systems and \nnFull inference systems to alleviate the need to evaluate an expensive predicate on all samples. The key method used in approximate filtering is to introduce an inexpensive predicate that approximates the true, expensive predicate. Due to the inexpensive nature of this approximate predicate, the approximate predicate can be evaluated on many more samples than the expensive one. Approximate filtering then works by evaluating the inexpensive approximate predicate on all samples to filter out those that one is confident will not satisfy the expensive predicate, and applying the true, expensive predicate on only those samples that satisfy the inexpensive predicate. Thus, approximate filtering accelerates search by reducing the number of samples on which the expensive predicate must be applied. This approach has been widely used in analytics systems through techniques such as model specialization~\cite{kang2017noscope}. 

\subsubsection{Applying approximate filtering to \nas} \label{sec:accel:opportunity:filtering_nas}
Our aim is to apply approximate filtering to reduce the number of \nnsFull that must undergo full accuracy evaluation (\ie $\symEvalFnAcc$) in sample-based \nas. To do so, we introduce an approximate evaluation function $\symEvalFnProxy$ that has an evaluation time $\symEvalTimeProxy \ll \symEvalTimeAcc$. 
With such an approximate evaluation function in place, the approximate-filtering-augmented \nas procedure is as follows: (1)~find the \pf ($\symFrontierProxy$) parameterized by \throughput and this approximate evaluation function by evaluating every \nnFull (or most) from the \searchspace in terms of \throughput and the approximate function; (2)~evaluate accuracy on each \nnFull in $\symFrontierProxy$ to obtain an approximate accuracy-\throughput \pf $\symFrontierAccApprox$.

\subsection{Utilization-based approximate filtering} \label{sec:accel:technique}
An approach to approximate filtering for sample-based \nas requires the use of an approximate evaluation function that (1)~correlates closely with accuracy (so as to preserve the quality of the final \pf found), (2)~is far less expensive to evaluate than accuracy, and (3)~significantly reduces the number of samples that require full accuracy evaluation. 

Based on our analysis of the \NATSbench search space in \Section\ref{sec:nas:survey}, we question whether \hardwareutilization could be a suitable approximate evaluation metric. Indeed, \hardwareutilization satisfies the first two criteria listed above: (1)~As shown in \Section\ref{sec:nas:survey}, \hardwareutilization correlates to accuracy for this search space; (2)~\Hardwareutilization is significantly less expensive to measure than accuracy, as it involves multiplying throughput by \FLOP count.

To analyze the third criterion, we count the number of networks in the \pf between throughput and \hardwareutilization, compared to the size of the total search space. We find that this \pf contains 144 networks, whereas the \NATSbench size search space contains 32767 networks as a whole. This illustrates that leveraging \hardwareutilization as an approximate evaluation function could significantly reduce the number of networks that require full accuracy evaluation, as only those within this \pf would need to be evaluated.

\paragraph{Why not use \FLOP count?}
A natural question that may arise is whether an alternative approximate evaluation function besides \hardwareutilization may suffice, such as using the \FLOP count of the \cnn. After all, \FLOP count typically does correlate with the accuracy of a \cnn. However, we find that using \FLOP count as an approximate evaluation function leads to an approximate \pf for the \NATSbench size search space with 255 evaluation points, which is significantly more than the 144 when using \hardwareutilization. Therefore, we opt for using \hardwareutilization as an approximate evaluation function in this work.

\begin{figure}[!t]
	\centering
    \includegraphics[width=1.0\linewidth]{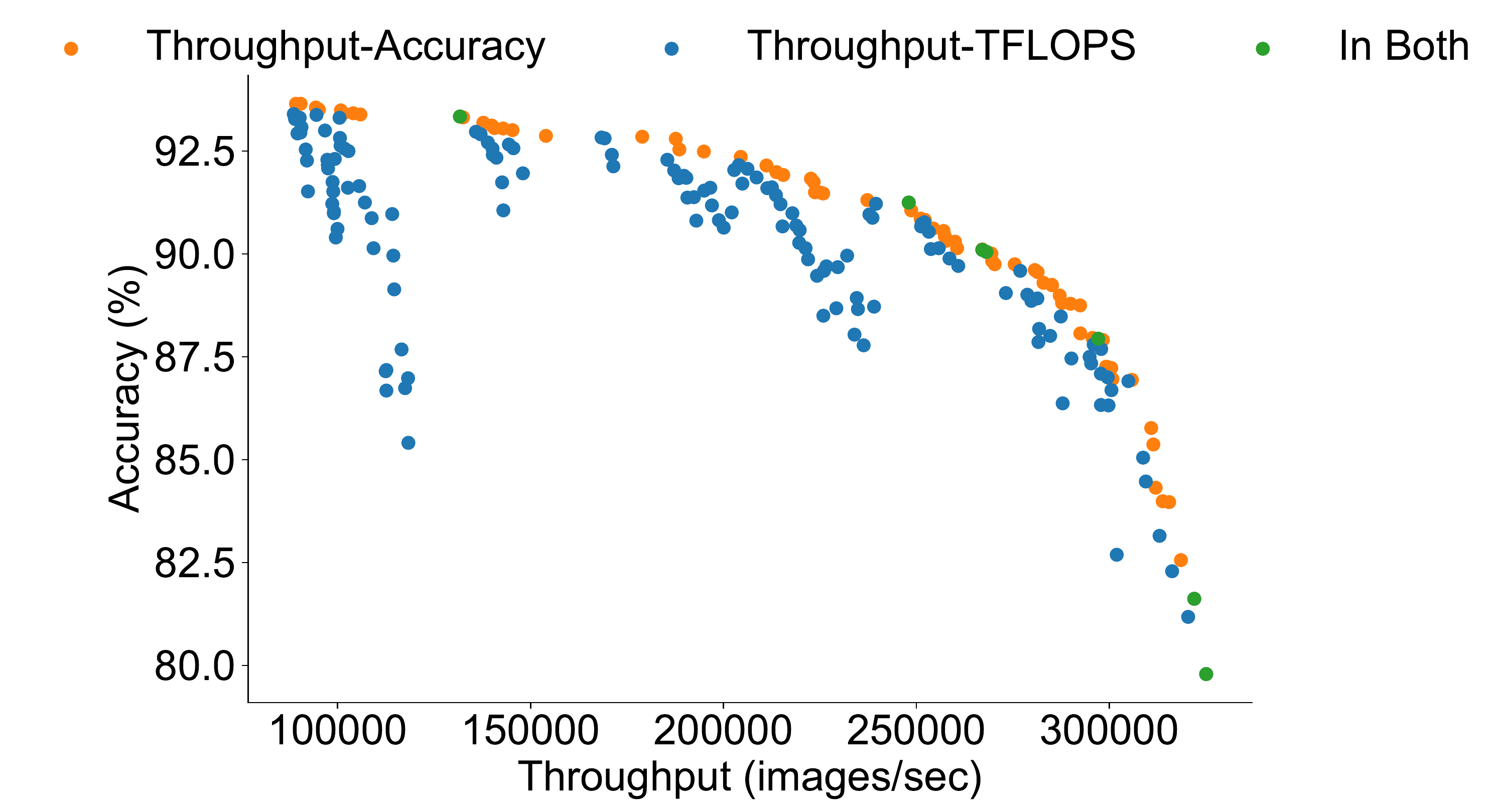}
    \caption{CNNs plotted in terms of their accuracy and throughput. Orange networks are those on the true throughput-accuracy \pf, blue networks are those on the throughput-utilization frontier, and green networks are those in both frontiers.}
    \label{fig:nas:approx}
\end{figure}

\paragraph{How close to the true \pf is the approximate \pf?}
Recall that the first step in leveraging approximate filtering for \nas is to generate an approximate \pf from which we will sample networks to evaluate accuracy. In order for the final results of this approximate search to maintain high fidelity to the original search, it is important that this approximate \pf contain many points close to those on the desired \pf. \Figure\ref{fig:nas:approx} compares the final throughput and accuracy of \cnns on this approximate frontier to those on the true throughput-accuracy frontier. We observe that the approximate frontier contains many networks that are very close to those on the true frontier, but also contains a number of spurious networks that are far from the true frontier. These spurious points will be eliminated by the second step, which performs accuracy evaluation on this approximate frontier, leading to an overall \pf close to the true frontier. Thus, we conclude that the overall approximate search procedure proposed will obtain a \pf close to that desired.

\subsection{Evaluating utilization-based approximate filtering} \label{sec:nas:accel:eval}
We now evaluate whether this approach of approximate filtering can reach better final \pfs than performing traditional \nas under the same time budget.

In doing so, we adopt the search simulation infrastructure developed in \NATSbench. We compare our proposed approach to using sampling-based \nas in which the sampling function is learned via REINFORCE with the reward function given by \Equation\ref{eqn:nas_eval} with $\symTputGoal= 175000$ and $\symExpTput=0.07$, borrowing this from~\citet{tan2019mnasnet}. We run each approach for 110000 timestep in the \NATSbench search simulation procedure: this is sufficient time for the approximate search technique to both evaluate the approximate metric on all points in the search space and to perform accuracy evaluation on the approximate \pf.

\begin{figure}
    \centering
    \includegraphics[width=0.8\linewidth]{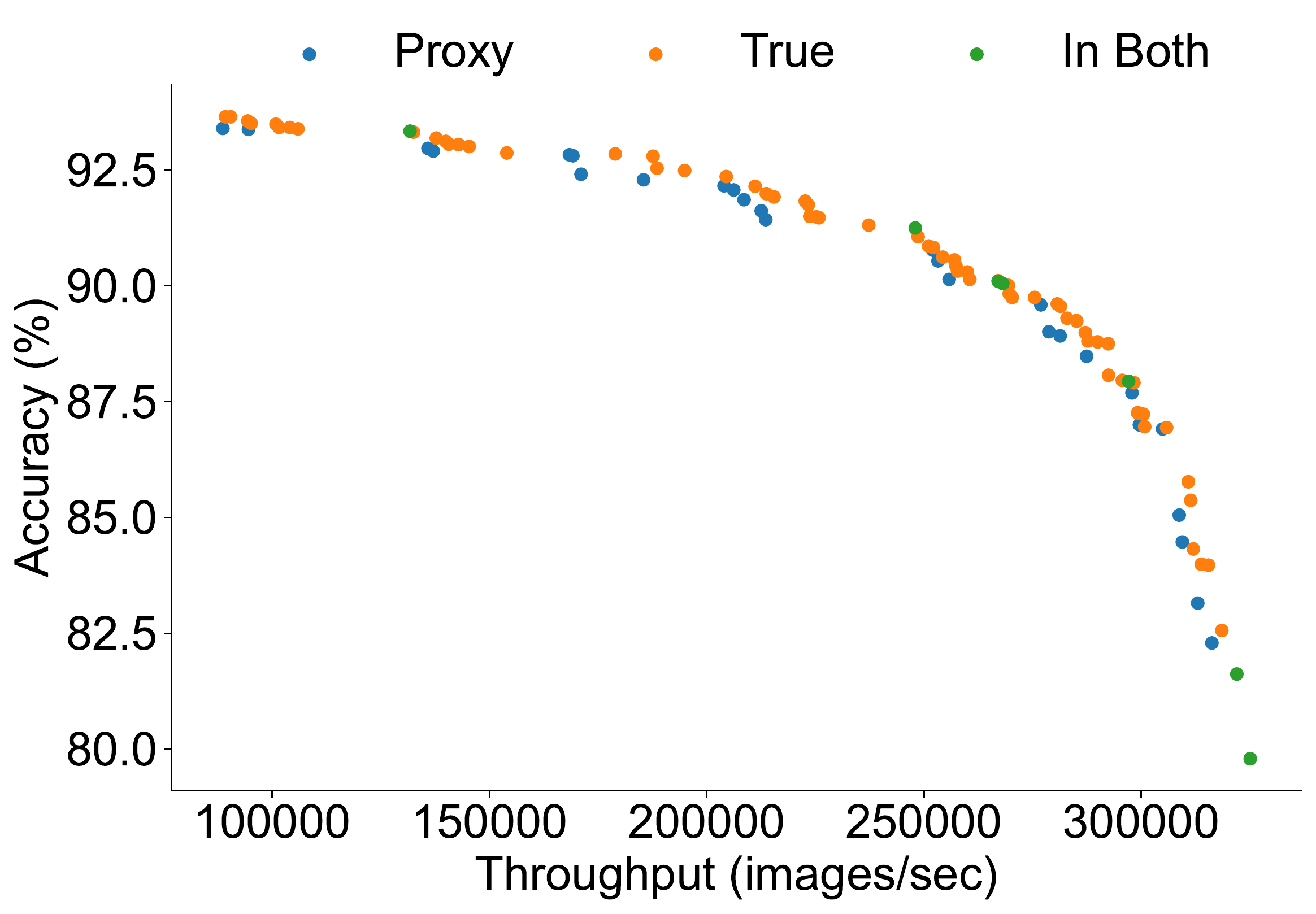}
    \caption{Accuracy-throughput \pf found by leveraging \hardwareutilization as an approximate evaluation function, via the algorithm described in \Section\ref{sec:accel:opportunity:filtering_nas}. ``Proxy'' refers to points returned by this approximate search. ``True'' refers to points returned by the traditional throughput-accuracy search. ``In Both'' refers to points that lie in each set of returned values.}
    \label{fig:nas:approx_results}
\end{figure}

\Figure\ref{fig:nas:approx_results} shows the \pf returned by the approximate filtering search procedure compared to the true accuracy-throughput \pf for this search space. The \pf returned by approximate search is similar to the true \pf: while the approximate \pf does contain fewer points than the true \pf, those points which are returned lie close to the true \pf. The small differences in accuracy and throughput between points on the approximate frontier and those on the true frontier may be indiscernible for many applications.

\begin{figure*}[!t]
	\centering
    \begin{subfigure}{0.3\textwidth}
        \includegraphics[width=\textwidth]{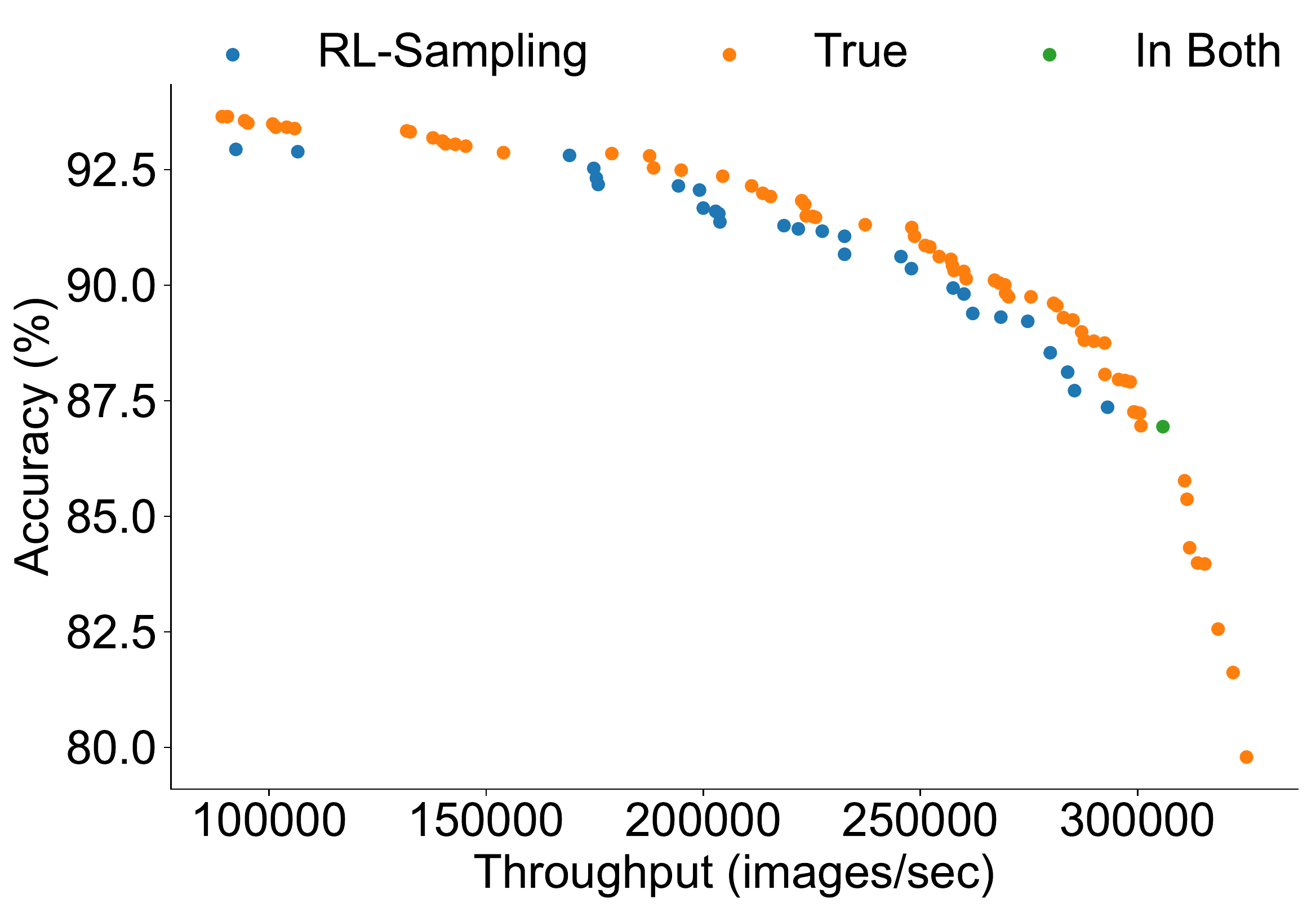}
    	\caption{Seed 1000}
    	\label{fig:rl:all:01000}
    \end{subfigure}
    \hfill
    \begin{subfigure}{0.3\textwidth}
        \includegraphics[width=\textwidth]{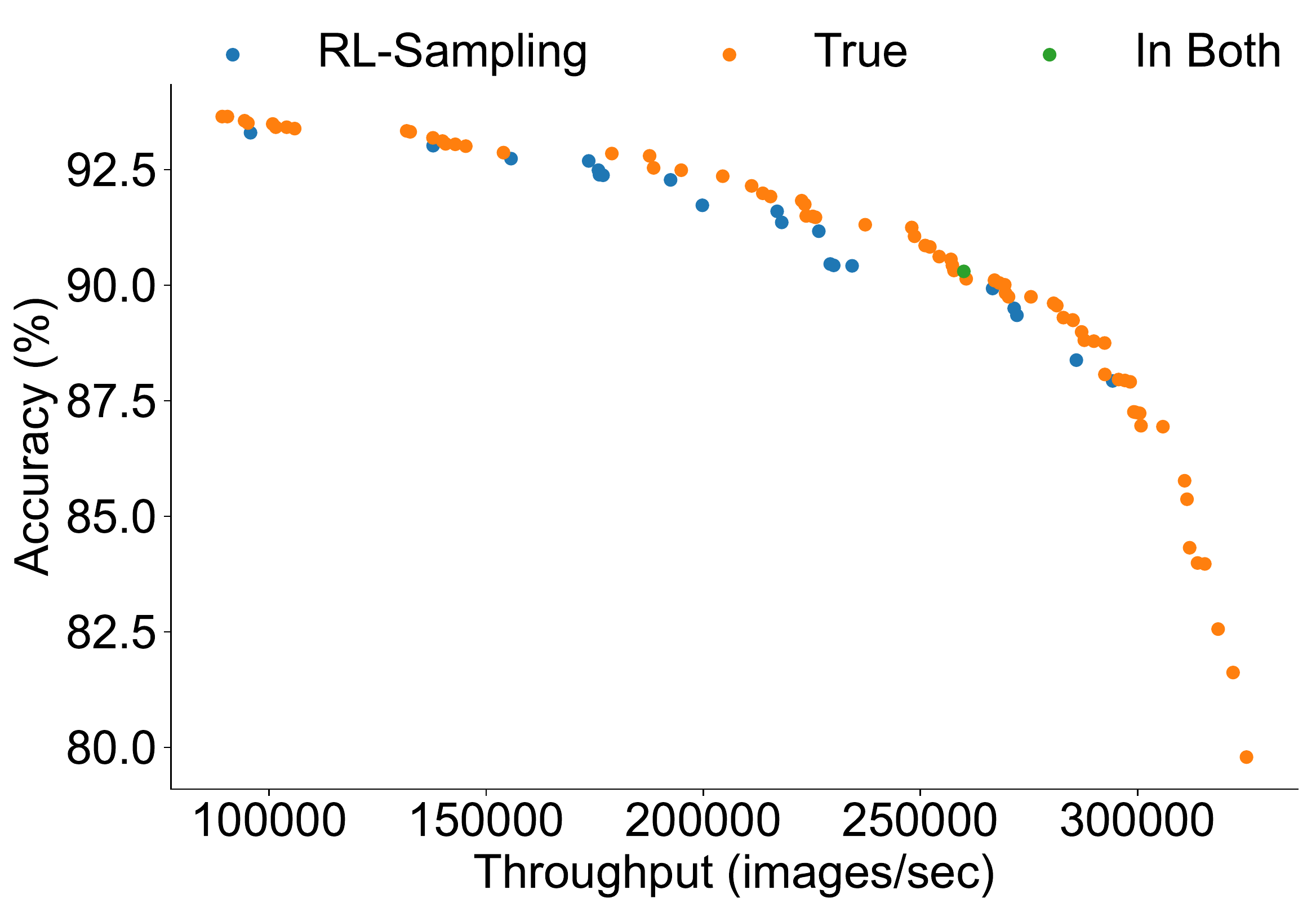}
    	\caption{Seed 2000}
    	\label{fig:rl:all:02000}
    \end{subfigure}
    \hfill
    \begin{subfigure}{0.3\textwidth}
        \includegraphics[width=\textwidth]{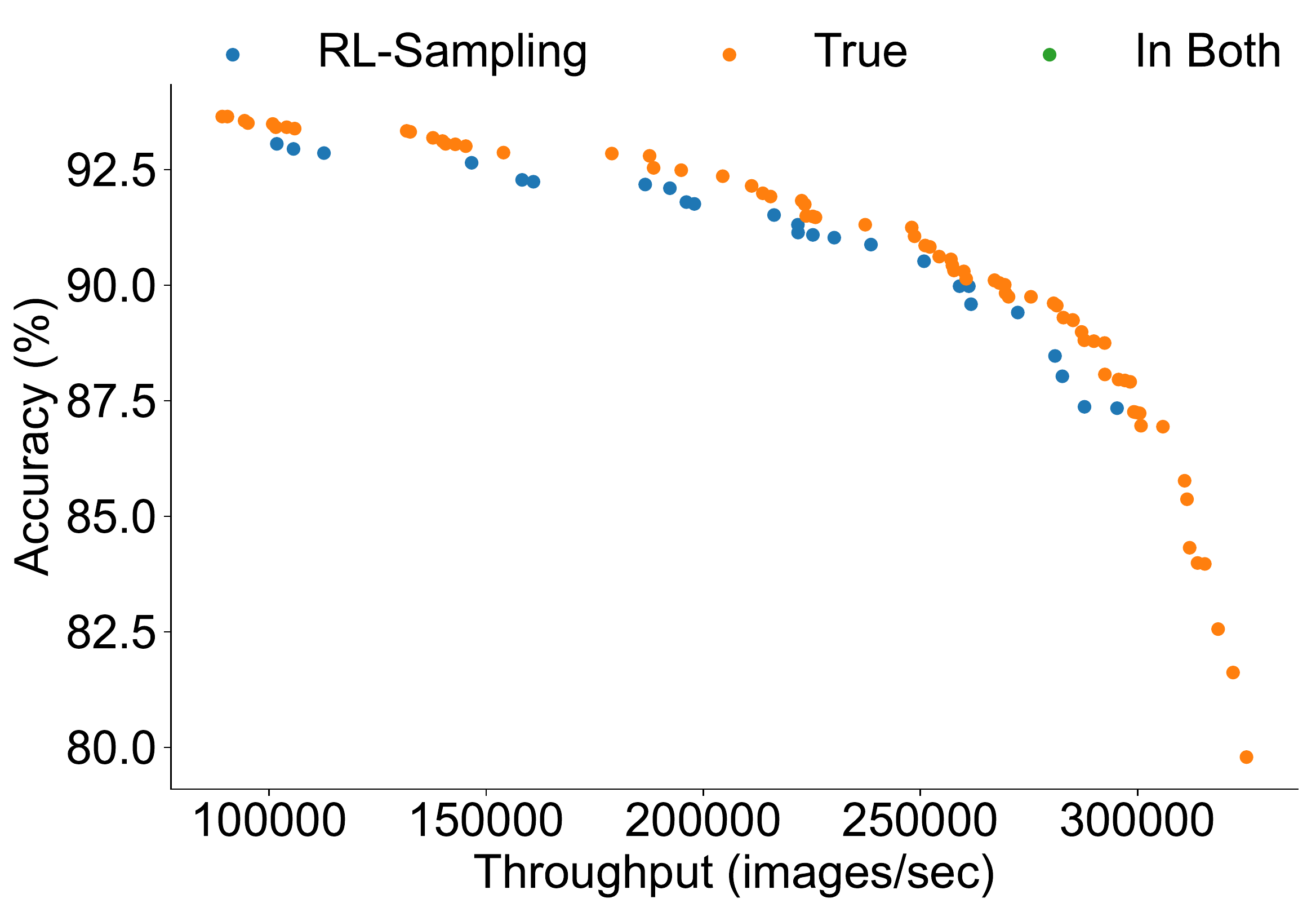}
    	\caption{Seed 3000}
    	\label{fig:rl:all:03000}
    \end{subfigure}
    \hfill
    \begin{subfigure}{0.3\textwidth}
        \includegraphics[width=\textwidth]{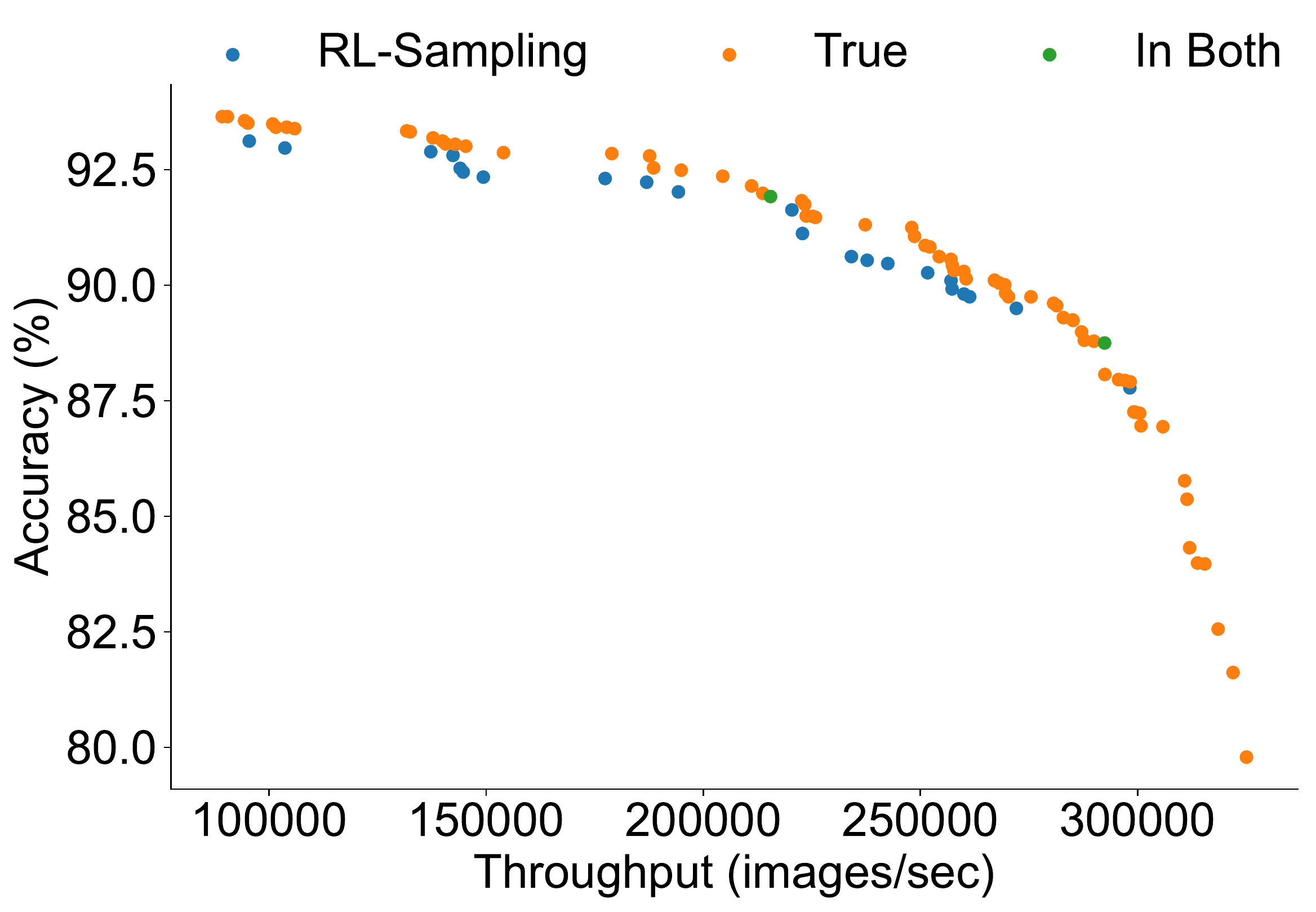}
    	\caption{Seed 4000}
    	\label{fig:rl:all:04000}
    \end{subfigure}
    \hfill
    \begin{subfigure}{0.3\textwidth}
        \includegraphics[width=\textwidth]{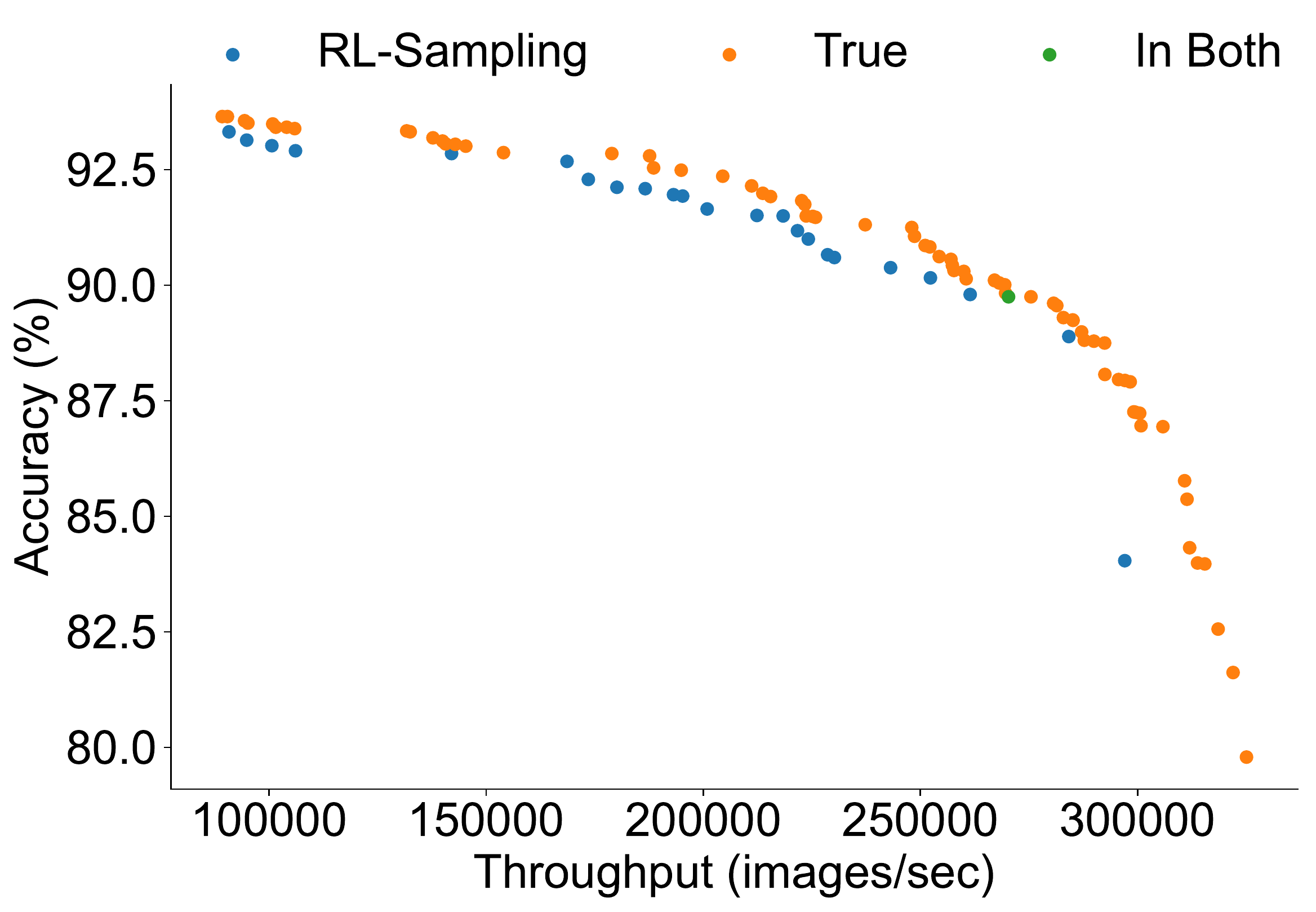}
    	\caption{Seed 5000}
    	\label{fig:rl:all:05000}
    \end{subfigure}
    \hfill
    \begin{subfigure}{0.3\textwidth}
        \includegraphics[width=\textwidth]{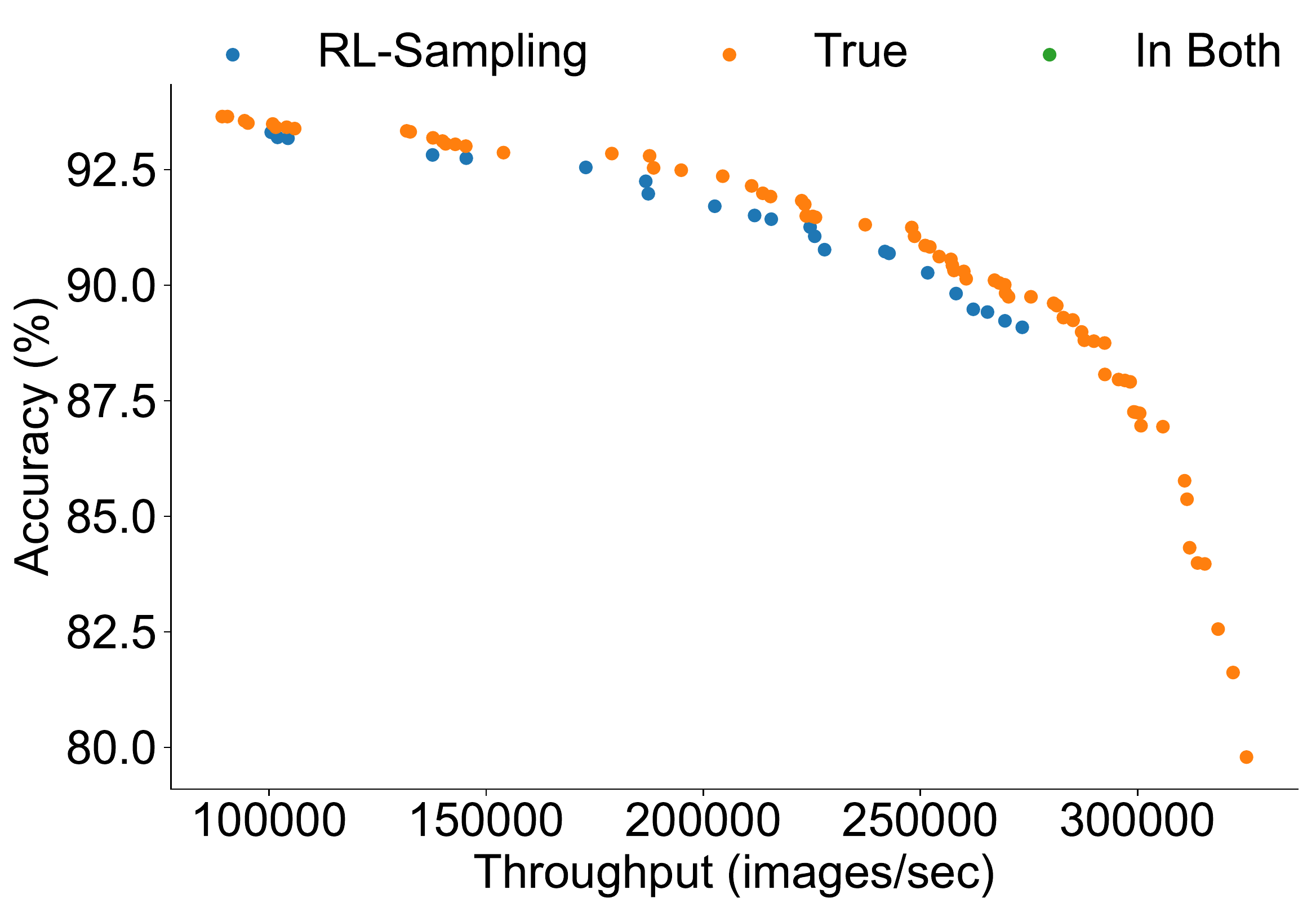}
    	\caption{Seed 6000}
    	\label{fig:rl:all:06000}
    \end{subfigure}
    \hfill
    \begin{subfigure}{0.3\textwidth}
        \includegraphics[width=\textwidth]{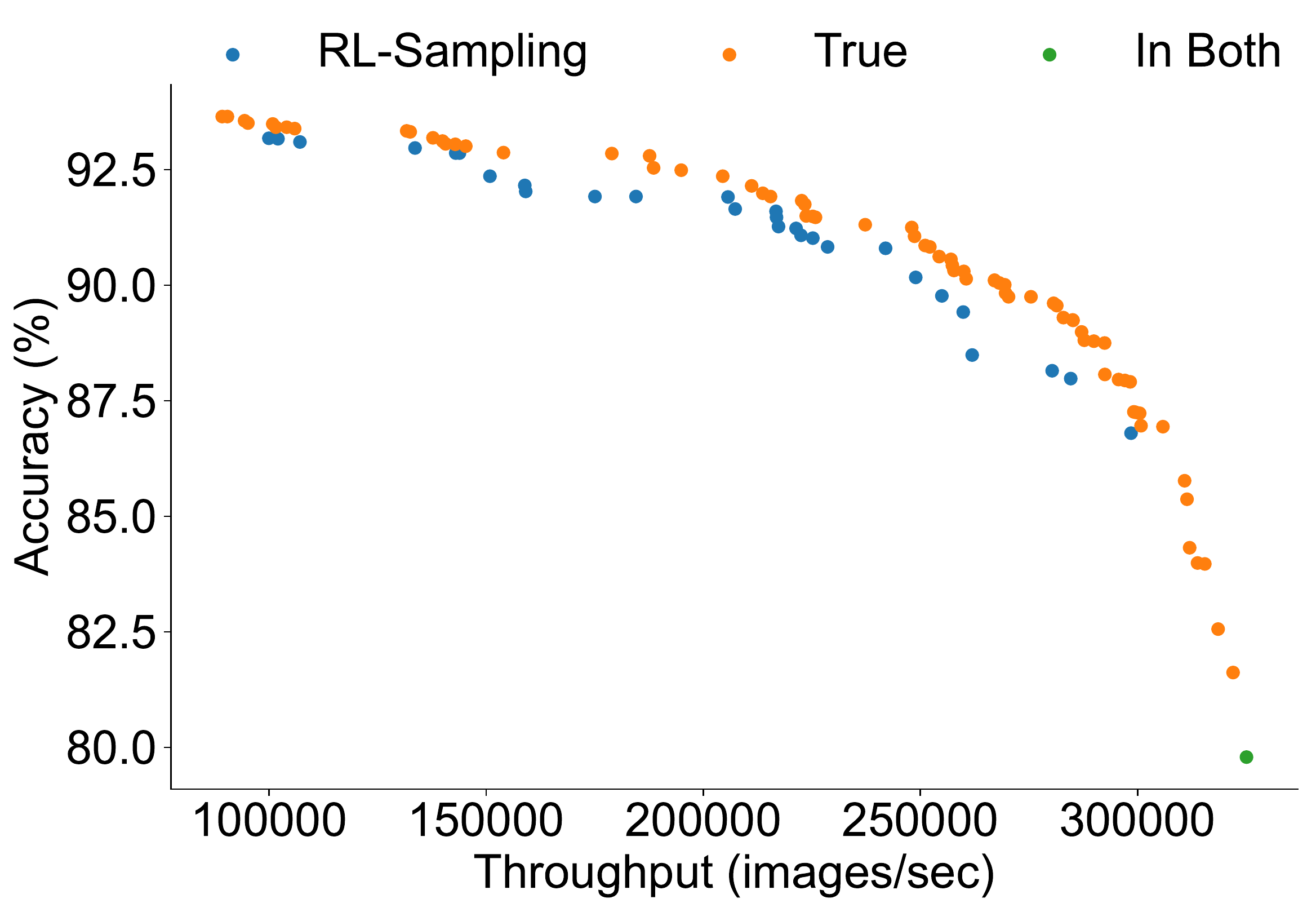}
    	\caption{Seed 7000}
    	\label{fig:rl:all:07000}
    \end{subfigure}
    \hfill
    \begin{subfigure}{0.3\textwidth}
        \includegraphics[width=\textwidth]{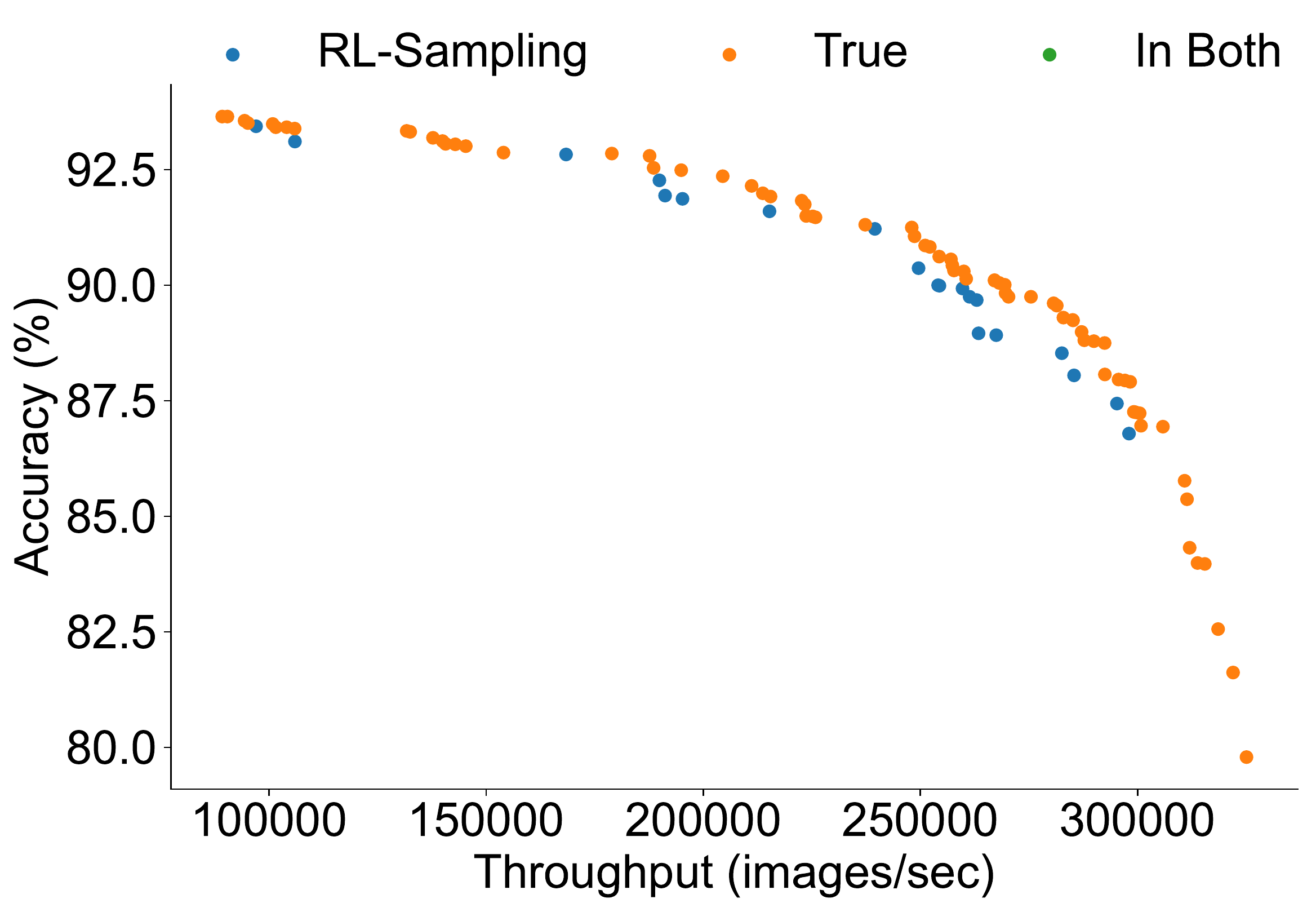}
    	\caption{Seed 8000}
    	\label{fig:rl:all:08000}
    \end{subfigure}
    \hfill
    \begin{subfigure}{0.3\textwidth}
        \includegraphics[width=\textwidth]{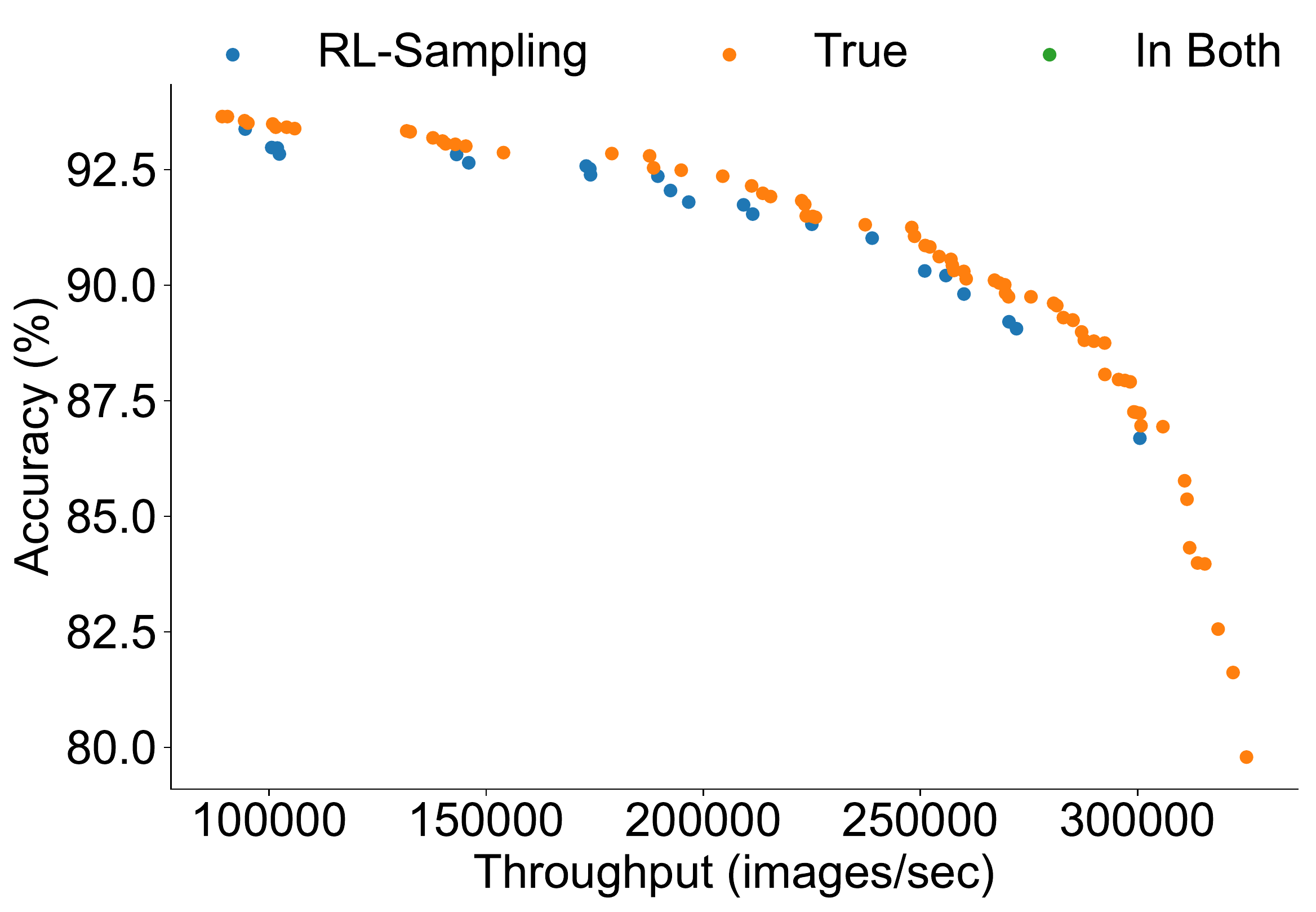}
    	\caption{Seed 9000}
    	\label{fig:rl:all:09000}
    \end{subfigure}
    \caption{Accuracy-throughput \pf achieved by reinforcement-learning-based \nas (blue) with nine different beginning random seeds, compared to the true \pf over the \NATSbench size search space (orange). ``In Both'' refers to points that lie in each set of returned values.}
    \label{fig:rl:all}
\end{figure*}

\Figure\ref{fig:rl:all} plots the \pfs returned by using reinforcement learning in \nas as described above, compared to the true \pf. We show this for nine different random seeds, as this technique is only able to sample a subset of the networks from the search space, and thus depends on the initial network one starts with. We generally find that the \pfs returned via this reinforcement-learning-based approach are also close to the true \pf. Some seeds show exceptions: for example, in seed 7000, the returned \pf is noticeably worse than the true \pf.

The results between reinforcement-learning-based search and leveraging the approximate filtering technique proposed above are mostly similar. This indicates that, while leveraging approximate filtering may be promising, it does not yield significant improvements in the returned \pf, at least for the \NATSbench size search space.

While the evaluation results above have not yielded significant benefits in terms of \nas search time or the quality of the networks returned from search, leveraging \hardwareutilization for approximate filtering within \nas may still raise benefits not revealed above. In particular, in many setups, evaluating accuracy may require a distributed training setup. This setup not only occupies significant cluster compute resources (\eg GPUs), but also consumes considerable network traffic for distributed training and storage system bandwidth for data loading~\cite{mohan2021analyzing}. In contrast, evaluating \hardwareutilization requires a single GPU, and does not consume any network or storage bandwidth. This reduces contention for these often-shared resources in datacenter and cloud settings. Thus, leveraging approximate filtering in \nas could make \nas more resource efficient.

\section{Conclusion}
This work has explored the \hardwareutilization of high-throughput \cnn inference as well as opportunities to leverage \hardwareutilization to improve \nas. We have evaluated the \hardwareutilization of popular \cnns, showing that these \cnns exhibit a wide range of \hardwareutilization. Many of these \cnns exhibit promising potential for improving \hardwareutilization, opening opportunities for future work to design \cnns that better utilize GPUs or system-level techniques to increase \hardwareutilization. Based on correlations between \hardwareutilization and accuracy, we next investigated the potential of using \hardwareutilization as a lightweight approximate evaluation metric in place of accuracy in \nas. We adapt the well-studied technique of approximate filtering from data systems literature to \nas. While the results from this preliminary study did not yield significant improvements in \nas search time or quality, we believe that future \nas techniques could benefit from considering leveraging \hardwareutilization in place of accuracy during \nas.

Through this exploration, this paper illustrates that \hardwareutilization is an important metric to consider both in terms of improving operational efficiency and \cnn throughput, as well as for potentially improving \nas algorithms.

\bibliography{references}

\begin{thebibliography}{34}
\providecommand{\natexlab}[1]{#1}
\providecommand{\url}[1]{\texttt{#1}}
\expandafter\ifx\csname urlstyle\endcsname\relax
  \providecommand{\doi}[1]{doi: #1}\else
  \providecommand{\doi}{doi: \begingroup \urlstyle{rm}\Url}\fi

\bibitem[Cai et~al.(2019)Cai, Gan, Wang, Zhang, and Han]{cai2019once}
Cai, H., Gan, C., Wang, T., Zhang, Z., and Han, S.
\newblock {Once-for-All: Train One Network and Specialize it for Efficient
  Deployment}.
\newblock In \emph{International Conference on Learning Representations ({ICLR}
  19)}, 2019.

\bibitem[Chen et~al.(2018)Chen, Moreau, Jiang, Zheng, Yan, Shen, Cowan, Wang,
  Hu, Ceze, Guestrin, and Krishnamurthy]{chen2018tvm}
Chen, T., Moreau, T., Jiang, Z., Zheng, L., Yan, E., Shen, H., Cowan, M., Wang,
  L., Hu, Y., Ceze, L., Guestrin, C., and Krishnamurthy, A.
\newblock {TVM}: An automated end-to-end optimizing compiler for deep learning.
\newblock In \emph{13th {USENIX} Symposium on Operating Systems Design and
  Implementation ({OSDI} 18)}, 2018.

\bibitem[Dong et~al.(2021)Dong, Liu, Musial, and Gabrys]{dong2021nats}
Dong, X., Liu, L., Musial, K., and Gabrys, B.
\newblock {NATS-Bench: Benchmarking NAS Algorithms for Architecture Topology
  and Size}.
\newblock \emph{IEEE Transactions on Pattern Analysis and Machine Intelligence
  ({IEEE TPAMI})}, 2021.

\bibitem[Elsken et~al.(2019)Elsken, Metzen, and Hutter]{elsken2019neural}
Elsken, T., Metzen, J.~H., and Hutter, F.
\newblock {Neural Architecture Search: A Survey}.
\newblock \emph{Journal of Machine Learning Research ({JMLR})}, 20\penalty0
  (55):\penalty0 1--21, 2019.

\bibitem[Gu et~al.(2019)Gu, Chowdhury, Shin, Zhu, Jeon, Qian, Liu, and
  Guo]{gu2019tiresias}
Gu, J., Chowdhury, M., Shin, K.~G., Zhu, Y., Jeon, M., Qian, J., Liu, H., and
  Guo, C.
\newblock {Tiresias: A GPU Cluster Manager for Distributed Deep Learning}.
\newblock In \emph{16th USENIX Symposium on Networked Systems Design and
  Implementation (NSDI 19)}, 2019.

\bibitem[Jain et~al.(2018)Jain, Mo, Jain, Subbaraj, Durrani, Tumanov, Gonzalez,
  and Stoica]{jain2018dynamic}
Jain, P., Mo, X., Jain, A., Subbaraj, H., Durrani, R.~S., Tumanov, A.,
  Gonzalez, J., and Stoica, I.
\newblock {Dynamic Space-Time Scheduling for GPU Inference}.
\newblock In \emph{NeurIPS Workshop on Systems for Machine Learning}, 2018.

\bibitem[Jia et~al.(2018)Jia, Maggioni, Staiger, and
  Scarpazza]{jia2018dissecting}
Jia, Z., Maggioni, M., Staiger, B., and Scarpazza, D.~P.
\newblock {Dissecting the NVIDIA Volta GPU Architecture via Microbenchmarking}.
\newblock \emph{arXiv preprint arXiv:1804.06826}, 2018.

\bibitem[Kang et~al.(2017)Kang, Emmons, Abuzaid, Bailis, and
  Zaharia]{kang2017noscope}
Kang, D., Emmons, J., Abuzaid, F., Bailis, P., and Zaharia, M.
\newblock {NoScope: Optimizing Neural Network Queries over Video at Scale}.
\newblock \emph{Proceedings of the VLDB Endowment}, 10\penalty0 (11):\penalty0
  1586--1597, 2017.

\bibitem[Kosaian et~al.(2021)Kosaian, Phanishayee, Philipose, Dey, and
  Rashmi]{kosaian2021boosting}
Kosaian, J., Phanishayee, A., Philipose, M., Dey, D., and Rashmi, K.~V.
\newblock {Boosting the Throughput and Accelerator Utilization of Specialized
  CNN Inference Beyond Increasing Batch Size}.
\newblock In \emph{Proceedings of the 38th International Conference on Machine
  Learning ({ICML} 21)}, 2021.

\bibitem[Li \& Talwalkar(2020)Li and Talwalkar]{li2020random}
Li, L. and Talwalkar, A.
\newblock {Random Search and Reproducibility for Neural Architecture Search}.
\newblock In \emph{Uncertainty in Artificial Intelligence ({UAI} 20)}, 2020.

\bibitem[Liu et~al.(2018)Liu, Simonyan, and Yang]{liu2018darts}
Liu, H., Simonyan, K., and Yang, Y.
\newblock {DARTS: Differentiable Architecture Search}.
\newblock In \emph{International Conference on Learning Representations ({ICLR}
  18)}, 2018.

\bibitem[Mohan et~al.(2021)Mohan, Phanishayee, Raniwala, and
  Chidambaram]{mohan2021analyzing}
Mohan, J., Phanishayee, A., Raniwala, A., and Chidambaram, V.
\newblock {Analyzing and Mitigating Data Stalls in {DNN} Training}.
\newblock \emph{Proceedings of the {VLDB} Endowment}, 14\penalty0 (5):\penalty0
  771--784, 2021.

\bibitem[Molchanov et~al.(2021)Molchanov, Hall, Yin, Kautz, Fusi, and
  Vahdat]{molchanov2021hant}
Molchanov, P., Hall, J., Yin, H., Kautz, J., Fusi, N., and Vahdat, A.
\newblock {HANT: Hardware-Aware Network Transformation}.
\newblock \emph{arXiv preprint arXiv:2107.10624}, 2021.

\bibitem[Narayanan et~al.(2018)Narayanan, Santhanam, Phanishayee, and
  Zaharia]{narayanan2018accelerating}
Narayanan, D., Santhanam, K., Phanishayee, A., and Zaharia, M.
\newblock {Accelerating Deep Learning Workloads Through Efficient Multi-Model
  Execution}.
\newblock In \emph{NeurIPS Workshop on Systems for Machine Learning}, 2018.

\bibitem[{NVIDIA}({\natexlab{a}})]{nvidia-a100}
{NVIDIA}.
\newblock {NVIDIA A100 GPU}.
  \url{https://www.nvidia.com/en-us/data-center/a100/}, {\natexlab{a}}.
\newblock Last accessed 23 November 2021.

\bibitem[{NVIDIA}({\natexlab{b}})]{nvidia-tensorcores}
{NVIDIA}.
\newblock {NVIDIA Tensor Cores
  \url{https://www.nvidia.com/en-us/data-center/tensor-cores/}},
  {\natexlab{b}}.
\newblock Last accessed 23 August 2021.

\bibitem[{NVIDIA}({\natexlab{c}})]{nvidia-tensorrt}
{NVIDIA}.
\newblock {NVIDIA TensorRT. \url{https://developer.nvidia.com/tensorrt}},
  {\natexlab{c}}.
\newblock Last accessed 23 August 2021.

\bibitem[{NVIDIA}(2017)]{nvidia-v100}
{NVIDIA}.
\newblock {NVIDIA Tesla V100 GPU Architecture}.
\newblock Technical Report WP-08608-001\_v1.1, 2017.

\bibitem[{NVIDIA}(2018)]{nvidia-t4}
{NVIDIA}.
\newblock {NVIDIA Turing GPU Architecture}.
\newblock Technical Report WP-09183-001\_v01, 2018.

\bibitem[Pham et~al.(2018)Pham, Guan, Zoph, Le, and Dean]{pham2018efficient}
Pham, H., Guan, M., Zoph, B., Le, Q., and Dean, J.
\newblock {Efficient Neural Architecture Search via Parameters Sharing}.
\newblock In \emph{Proceedings of the 35th International Conference on Machine
  Learning ({ICML} 18)}, 2018.

\bibitem[Real et~al.(2019)Real, Aggarwal, Huang, and Le]{real2019regularized}
Real, E., Aggarwal, A., Huang, Y., and Le, Q.~V.
\newblock {Regularized Evolution for Image Classifier Architecture Search}.
\newblock In \emph{Proceedings of the {AAAI} Conference on Artificial
  Intelligence ({AAAI} 19)}, 2019.

\bibitem[Redmon et~al.(2016)Redmon, Divvala, Girshick, and
  Farhadi]{redmon2016you}
Redmon, J., Divvala, S., Girshick, R., and Farhadi, A.
\newblock {You Only Look Once: Unified, Real-Time Object Detection}.
\newblock In \emph{Proceedings of the IEEE Conference on Computer Vision and
  Pattern Recognition ({CVPR} 16)}, 2016.

\bibitem[Ridnik et~al.(2021)Ridnik, Lawen, Noy, Ben~Baruch, Sharir, and
  Friedman]{ridnik2021tresnet}
Ridnik, T., Lawen, H., Noy, A., Ben~Baruch, E., Sharir, G., and Friedman, I.
\newblock {Tresnet: High Performance GPU-Dedicated Architecture}.
\newblock In \emph{Proceedings of the IEEE/CVF Winter Conference on
  Applications of Computer Vision ({WACV} 21)}, 2021.

\bibitem[Russakovsky et~al.(2015)Russakovsky, Deng, Su, Krause, Satheesh, Ma,
  Huang, Karpathy, Khosla, Bernstein, Berg, and
  Fei-Fei]{russakovsky2015imagenet}
Russakovsky, O., Deng, J., Su, H., Krause, J., Satheesh, S., Ma, S., Huang, Z.,
  Karpathy, A., Khosla, A., Bernstein, M., Berg, A.~C., and Fei-Fei, L.
\newblock {ImageNet Large Scale Visual Recognition Challenge}.
\newblock \emph{International Journal of Computer Vision (IJCV)}, 115\penalty0
  (3):\penalty0 211--252, 2015.
\newblock \doi{10.1007/s11263-015-0816-y}.

\bibitem[Tan \& Le(2019)Tan and Le]{tan2019efficientnet}
Tan, M. and Le, Q.
\newblock {{E}fficient{N}et: Rethinking Model Scaling for Convolutional Neural
  Networks}.
\newblock In \emph{Proceedings of the 36th International Conference on Machine
  Learning ({ICML} 19)}, 2019.

\bibitem[Tan et~al.(2019)Tan, Chen, Pang, Vasudevan, Sandler, Howard, and
  Le]{tan2019mnasnet}
Tan, M., Chen, B., Pang, R., Vasudevan, V., Sandler, M., Howard, A., and Le,
  Q.~V.
\newblock {MnasNet: Platform-Aware Neural Architecture Search for Mobile}.
\newblock In \emph{Proceedings of the IEEE/CVF Conference on Computer Vision
  and Pattern Recognition ({CVPR} 19)}, 2019.

\bibitem[Wang et~al.(2021)Wang, Yang, Zheng, Li, and
  Pekhimenko]{wang2021horizontally}
Wang, S., Yang, P., Zheng, Y., Li, X., and Pekhimenko, G.
\newblock {Horizontally Fused Training Array: An Effective Hardware Utilization
  Squeezer for Training Novel Deep Learning Models}.
\newblock \emph{The Fourth Conference on Machine Learning and Systems ({MLSys}
  21)}, 2021.

\bibitem[Wightman(2019)]{wightman2019timm}
Wightman, R.
\newblock Pytorch image models.
\newblock \url{https://github.com/rwightman/pytorch-image-models}, 2019.

\bibitem[Williams et~al.(2009)Williams, Waterman, and
  Patterson]{williams2009roofline}
Williams, S., Waterman, A., and Patterson, D.
\newblock {Roofline: an Insightful Visual Performance Model for Multicore
  Architectures}.
\newblock \emph{Communications of the ACM}, 52\penalty0 (4):\penalty0 65--76,
  2009.

\bibitem[Wu et~al.(2019)Wu, Dai, Zhang, Wang, Sun, Wu, Tian, Vajda, Jia, and
  Keutzer]{wu2019fbnet}
Wu, B., Dai, X., Zhang, P., Wang, Y., Sun, F., Wu, Y., Tian, Y., Vajda, P.,
  Jia, Y., and Keutzer, K.
\newblock {FBNet: Hardware-Aware Efficient ConvNet Design via Differentiable
  Neural Architecture Search}.
\newblock In \emph{Proceedings of the IEEE/CVF Conference on Computer Vision
  and Pattern Recognition}, pp.\  10734--10742, 2019.

\bibitem[Xiao et~al.(2018)Xiao, Bhardwaj, Ramjee, Sivathanu, Kwatra, Han,
  Patel, Peng, Zhao, Zhang, et~al.]{xiao2018gandiva}
Xiao, W., Bhardwaj, R., Ramjee, R., Sivathanu, M., Kwatra, N., Han, Z., Patel,
  P., Peng, X., Zhao, H., Zhang, Q., et~al.
\newblock {Gandiva: Introspective Cluster Scheduling for Deep Learning}.
\newblock In \emph{13th USENIX Symposium on Operating Systems Design and
  Implementation (OSDI 18)}, 2018.

\bibitem[Yu \& Chowdhury(2020)Yu and Chowdhury]{yu2020salus}
Yu, P. and Chowdhury, M.
\newblock {Salus: Fine-Grained GPU Sharing Primitives for Deep Learning
  Applications}.
\newblock In \emph{The Third Conference on Machine Learning and Systems
  ({MLSys} 20)}, 2020.

\bibitem[Zhou et~al.(2018)Zhou, Ebrahimi, Ar{\i}k, Yu, Liu, and
  Diamos]{zhou2018resource}
Zhou, Y., Ebrahimi, S., Ar{\i}k, S.~{\"O}., Yu, H., Liu, H., and Diamos, G.
\newblock {Resource-Efficient Neural Architect}.
\newblock \emph{arXiv preprint arXiv:1806.07912}, 2018.

\bibitem[Zoph \& Le(2016)Zoph and Le]{zoph2016neural}
Zoph, B. and Le, Q.~V.
\newblock {Neural Architecture Search with Reinforcement Learning}.
\newblock \emph{arXiv preprint arXiv:1611.01578}, 2016.

\end{thebibliography}
\bibliographystyle{mlsys2022}

\begin{appendices}
\section{Results from surveyed \cnns} \label{app:survey}
Tables~\ref{table:results} and~\ref{table:results_continued} show the results from the survey of \cnns performed in \Section\ref{sec:survey}. We additionally depict with ``\symOn'' whether a \cnn is on a particular Pareto frontier.
\begin{table*}[]
    \centering
    \caption{Models evaluated in survey. Accuracy is in percentage and throughput is in images/sec.}
    \begin{tabular}{c|c|c|c|c|c|c}
     & & & & On Tput.-Acc. & On TFLOPS-Acc. & On Tput.-TFLOPS \\
     
    Name & Accuracy & Throughput & TFLOPs/sec & Frontier? & Frontier? & Frontier? \\
\hline
tresnet\_m & 83.08 & 8444.73 & 96.86 & \symOn & \symOn & \symOn \\
efficientnetv2\_m & 85.10 & 1039.29 & 38.39 & \symOn & \symOn & \symOff \\
seresnet152d & 84.36 & 1175.21 & 56.39 & \symOn & \symOn & \symOff \\
gernet\_s & 76.92 & 31344.26 & 46.65 & \symOn & \symOff & \symOn \\
regnetx\_004 & 72.40 & 39215.73 & 31.19 & \symOn & \symOff & \symOn \\
regnetx\_002 & 68.76 & 50893.82 & 20.26 & \symOn & \symOff & \symOn \\
efficientnetv2\_s & 83.90 & 2176.89 & 36.42 & \symOn & \symOff & \symOff \\
gernet\_m & 80.73 & 12411.49 & 74.58 & \symOn & \symOff & \symOff \\
ofanet-v100\_@6ms & 73.00 & 32903.61 & 11.80 & \symOn & \symOff & \symOff \\
ofanet-v100\_@5ms & 71.00 & 42038.86 & 11.87 & \symOn & \symOff & \symOff \\
regnety\_002 & 70.25 & 42492.74 & 16.96 & \symOn & \symOff & \symOff \\
resnetrs270 & 84.43 & 535.07 & 54.49 & \symOff & \symOn & \symOff \\
resnetrs152 & 83.71 & 1169.59 & 56.72 & \symOff & \symOn & \symOff \\
resnet18 & 69.75 & 22409.05 & 81.30 & \symOff & \symOff & \symOn \\
resnetrs200 & 84.07 & 858.29 & 53.86 & \symOff & \symOff & \symOff \\
regnety\_032 & 82.72 & 2915.10 & 30.60 & \symOff & \symOff & \symOff \\
resnetrs101 & 82.29 & 2062.88 & 55.70 & \symOff & \symOff & \symOff \\
tresnet\_xl & 82.05 & 2552.27 & 77.41 & \symOff & \symOff & \symOff \\
tresnet\_l & 81.49 & 3411.52 & 74.21 & \symOff & \symOff & \symOff \\
wide\_resnet50 & 81.46 & 4002.17 & 91.23 & \symOff & \symOff & \symOff \\
gernet\_l & 81.35 & 8166.70 & 74.43 & \symOff & \symOff & \symOff \\
seresnext50\_32x4d & 81.27 & 3998.37 & 33.85 & \symOff & \symOff & \symOff \\
seresnext101\_32x4d & 80.90 & 2657.49 & 42.39 & \symOff & \symOff & \symOff \\
repvgg\_b3 & 80.49 & 1526.05 & 88.88 & \symOff & \symOff & \symOff \\
seresnet50 & 80.27 & 6059.46 & 49.59 & \symOff & \symOff & \symOff \\
repvgg\_b3g4 & 80.21 & 2082.79 & 74.35 & \symOff & \symOff & \symOff \\
dpn107 & 80.16 & 887.80 & 32.49 & \symOff & \symOff & \symOff \\
efficientnet\_b2 & 80.10 & 4518.30 & 9.83 & \symOff & \symOff & \symOff \\
cspresnext50 & 80.04 & 4911.09 & 30.20 & \symOff & \symOff & \symOff \\
dpn92 & 80.01 & 1677.89 & 21.82 & \symOff & \symOff & \symOff \\
ofanet-flops@595M & 80.00 & 7063.24 & 8.41 & \symOff & \symOff & \symOff \\
resnetrs50 & 79.89 & 5189.78 & 44.55 & \symOff & \symOff & \symOff \\
dpn131 & 79.82 & 1002.03 & 32.09 & \symOff & \symOff & \symOff \\
resnext50\_32x4d & 79.77 & 5011.84 & 42.41 & \symOff & \symOff & \symOff \\
regnety\_064 & 79.72 & 3313.35 & 42.11 & \symOff & \symOff & \symOff \\
resnext50d\_32x4d & 79.68 & 4755.81 & 42.52 & \symOff & \symOff & \symOff \\
dpn98 & 79.64 & 1435.61 & 33.51 & \symOff & \symOff & \symOff \\
ofanet-flops@482M & 79.60 & 9035.14 & 8.72 & \symOff & \symOff & \symOff \\
cspresnet50 & 79.57 & 6691.55 & 60.45 & \symOff & \symOff & \symOff \\
hrnet\_w64 & 79.47 & 1277.18 & 73.81 & \symOff & \symOff & \symOff \\
repvgg\_b2g4 & 79.37 & 2542.93 & 64.03 & \symOff & \symOff & \symOff \\
resnext101\_32x8d & 79.31 & 2265.54 & 74.37 & \symOff & \symOff & \symOff \\
hrnet\_w48 & 79.30 & 1612.77 & 55.73 & \symOff & \symOff & \symOff \\
regnety\_040 & 79.22 & 6893.49 & 54.78 & \symOff & \symOff & \symOff \\
dpn68b & 79.22 & 4644.92 & 21.61 & \symOff & \symOff & \symOff \\
regnetx\_080 & 79.19 & 4313.02 & 68.97 & \symOff & \symOff & \symOff \\
ofanet-flops@389M & 79.10 & 11343.95 & 8.84 & \symOff & \symOff & \symOff \\
efficientnet\_b1 & 79.10 & 6249.63 & 9.29 & \symOff & \symOff & \symOff
    \end{tabular}
    \label{table:results}
\end{table*}

\begin{table*}[]
    \caption{(Continued) Models evaluated in survey. Accuracy is in percentage and throughput is in images/sec.}
    \begin{tabular}{c|c|c|c|c|c|c}
     & & & & On Tput.-Acc. & On TFLOPS-Acc. & On Tput.-TFLOPS \\
    Name & Accuracy & Throughput & TFLOPs/sec & Frontier? & Frontier? & Frontier? \\
\hline
resnet50 & 79.04 & 8722.50 & 71.34 & \symOff & \symOff & \symOff \\
hrnet\_w40 & 78.92 & 1742.82 & 44.24 & \symOff & \symOff & \symOff \\
hrnet\_w44 & 78.90 & 1632.27 & 48.59 & \symOff & \symOff & \symOff \\
repvgg\_b2 & 78.79 & 1999.03 & 81.63 & \symOff & \symOff & \symOff \\
dla169 & 78.69 & 2610.92 & 60.35 & \symOff & \symOff & \symOff \\
dla102x & 78.51 & 3209.39 & 37.54 & \symOff & \symOff & \symOff \\
regnetx\_040 & 78.48 & 5317.33 & 42.16 & \symOff & \symOff & \symOff \\
dla60\_res2net & 78.46 & 4167.03 & 34.36 & \symOff & \symOff & \symOff \\
hrnet\_w32 & 78.45 & 2842.49 & 50.72 & \symOff & \symOff & \symOff \\
dla60\_res2next & 78.44 & 4035.40 & 27.91 & \symOff & \symOff & \symOff \\
seresnet101 & 78.38 & 3552.61 & 55.46 & \symOff & \symOff & \symOff \\
repvgg\_b1 & 78.37 & 3120.94 & 81.94 & \symOff & \symOff & \symOff \\
dla60x & 78.25 & 4394.39 & 30.91 & \symOff & \symOff & \symOff \\
hrnet\_w30 & 78.21 & 2745.27 & 44.51 & \symOff & \symOff & \symOff \\
regnetx\_032 & 78.17 & 7045.64 & 44.76 & \symOff & \symOff & \symOff \\
dla102 & 78.03 & 3795.49 & 54.39 & \symOff & \symOff & \symOff \\
seresnext26t\_32x4d & 77.99 & 6081.96 & 32.60 & \symOff & \symOff & \symOff \\
regnety\_016 & 77.86 & 7238.16 & 23.34 & \symOff & \symOff & \symOff \\
tv\_resnext50\_32x4d & 77.62 & 5015.94 & 42.44 & \symOff & \symOff & \symOff \\
seresnext26d\_32x4d & 77.60 & 6070.07 & 32.92 & \symOff & \symOff & \symOff \\
repvgg\_b1g4 & 77.59 & 4135.22 & 67.14 & \symOff & \symOff & \symOff \\
mixnet\_m & 77.26 & 4722.31 & 3.20 & \symOff & \symOff & \symOff \\
efficientnet\_b0 & 77.10 & 11612.13 & 8.96 & \symOff & \symOff & \symOff \\
dla60 & 77.03 & 5507.80 & 46.66 & \symOff & \symOff & \symOff \\
regnetx\_016 & 76.95 & 8326.46 & 26.69 & \symOff & \symOff & \symOff \\
hrnet\_w18 & 76.76 & 3198.62 & 27.40 & \symOff & \symOff & \symOff \\
repvgg\_a2 & 76.46 & 6444.23 & 73.28 & \symOff & \symOff & \symOff \\
dpn68 & 76.32 & 4105.41 & 19.10 & \symOff & \symOff & \symOff \\
regnety\_008 & 76.32 & 19957.41 & 31.82 & \symOff & \symOff & \symOff \\
ofanet-v100\_@11ms & 76.10 & 16811.99 & 11.83 & \symOff & \symOff & \symOff \\
mixnet\_s & 75.99 & 6812.00 & 3.26 & \symOff & \symOff & \symOff \\
ofanet-v100\_@9ms & 75.30 & 21176.05 & 13.25 & \symOff & \symOff & \symOff \\
regnety\_006 & 75.25 & 23534.66 & 28.30 & \symOff & \symOff & \symOff \\
repvgg\_b0 & 75.15 & 9027.21 & 61.33 & \symOff & \symOff & \symOff \\
hrnet\_w18\_small\_v2 & 75.11 & 6063.18 & 31.48 & \symOff & \symOff & \symOff \\
regnetx\_008 & 75.04 & 23851.86 & 38.15 & \symOff & \symOff & \symOff \\
seresnet34 & 74.81 & 10143.47 & 74.33 & \symOff & \symOff & \symOff \\
dla34 & 74.63 & 7573.85 & 46.35 & \symOff & \symOff & \symOff \\
regnety\_004 & 74.03 & 24692.65 & 19.85 & \symOff & \symOff & \symOff \\
regnetx\_006 & 73.85 & 18555.66 & 22.30 & \symOff & \symOff & \symOff \\
hrnet\_w18\_small & 72.34 & 11923.19 & 38.19 & \symOff & \symOff & \symOff \\
seresnet18 & 71.74 & 18435.65 & 66.89 & \symOff & \symOff & \symOff \\
dla60x\_c & 67.89 & 7120.45 & 8.28 & \symOff & \symOff & \symOff \\
dla46x\_c & 65.97 & 7567.59 & 8.06 & \symOff & \symOff & \symOff \\
dla46\_c & 64.87 & 12140.11 & 13.94 & \symOff & \symOff & \symOff 
    \end{tabular}
    \label{table:results_continued}
\end{table*}
\end{appendices}

\end{document}